\documentclass[10pt,onecolumn]{article}
\usepackage[T1]{fontenc}
\usepackage[utf8]{inputenc}
\usepackage[a4paper,margin=0.75in]{geometry}
\usepackage{microtype}
\usepackage{amsmath,amssymb}
\usepackage{booktabs}
\usepackage{graphicx}
\usepackage{float}
\usepackage[section]{placeins}
\usepackage[hidelinks]{hyperref}
\usepackage[numbers,sort&compress]{natbib}
\usepackage{authblk}
\usepackage{titlesec}
\usepackage[labelfont=bf,font=small]{caption}
\usepackage{enumitem}
\usepackage{xcolor}

\usepackage{etoolbox}
\newenvironment{revblock}{}{}

\graphicspath{{./}}
\titleformat{\section}{\normalfont\bfseries\large}{\thesection.}{0.5em}{}
\titleformat{\subsection}{\normalfont\bfseries\normalsize}{\thesubsection.}{0.5em}{}
\titleformat{\subsubsection}{\normalfont\itshape\normalsize}{\thesubsubsection.}{0.5em}{}
\titlespacing*{\section}{0pt}{1.2ex plus 0.3ex minus 0.2ex}{0.8ex}
\titlespacing*{\subsection}{0pt}{1.0ex plus 0.2ex minus 0.2ex}{0.5ex}
\titlespacing*{\subsubsection}{0pt}{0.8ex plus 0.2ex minus 0.2ex}{0.2ex}

\title{Geometric and Quantum Kernel Methods for Predicting Skeletal Muscle Outcomes in Chronic Obstructive Pulmonary Disease}
\author[1]{Azadeh Alavi*}
\author[2]{Hamidreza Khalili}
\author[2]{Stanley M. H.~Chan*}
\author[3]{Fatemeh Kouchmeshki}
\author[4]{Muhammad Usman}
\author[2]{Ross Vlahos}
\affil[1]{School of Computing Technologies, RMIT University, Melbourne 3000, Australia}
\affil[2]{School of Health \& Biomedical Sciences, STEM College, RMIT University, Melbourne 3000, Australia}
\affil[3]{Pattern Recognition Pty Ltd, Melbourne 3240, Australia}
\affil[4]{Data61, CSIRO, Clayton, VIC, Australia}
\date{}

\begin{document}
\maketitle
\noindent\textbf{Corresponding authors:} Azadeh Alavi azadeh.alavi@rmit.edu.au, Stanley M. H. Chan stanley.chan@rmit.edu.au

\vspace{0.5em}
\noindent These authors contributed equally to this work: Azadeh Alavi, Hamidreza Khalili and Stanley M.~H.~Chan.

{\small
\noindent\textbf{Abstract}\par
\begin{revblock}
Chronic obstructive pulmonary disease (COPD) affects hundreds of millions of people worldwide, and skeletal-muscle dysfunction is clinically important. Quantum machine learning is increasingly explored for biomedical prediction, but its value in small biomarker cohorts requires benchmarking against strong classical baselines. We analysed a cigarette-smoke COPD cohort of 213 animals with blood and bronchoalveolar-lavage biomarkers to predict tibialis anterior muscle weight, muscle quality, and force. We developed a kernel-geometric quantum hybrid method in which synthetic symmetric positive definite (SPD) references are mapped through a reproducing kernel Hilbert space, compressed using train-only random projection, normalised, and supplied to low-dimensional quantum regression circuits. We benchmarked this approach against classical ridge/kernel models, SPD relational representations, and quantum-kernel regression (QKR). All methods were evaluated using condition-stratified repeated cross-validation. The largest numerical improvement was observed for muscle weight, where the proposed method had the numerically lowest mean root mean squared error (RMSE), approximately 1.8\% below the best classical comparator; paired fold-level testing did not establish statistically significant superiority after Holm adjustment, but the endpoint is biologically meaningful. The method also had the numerically lowest mean RMSE for muscle quality. For force, biomarker-only Ridge performed best, suggesting a more linear endpoint structure.

\end{revblock}
}

\vspace{1em}

\section{Introduction}
Chronic obstructive pulmonary disease (COPD) is a chronic respiratory disorder characterised by persistent airflow limitation and progressive loss of lung function driven by abnormal inflammatory responses to noxious exposures, most commonly cigarette smoke \cite{ref1,ref2}. COPD is also a systemic disease: chronic inflammation, oxidative stress, hypoxia, and comorbid metabolic disturbance contribute to extrapulmonary consequences, including skeletal muscle dysfunction \cite{ref10,ref11,ref13}. This systemic involvement is clinically important because limb muscle dysfunction reduces exercise tolerance, impairs daily function, and predicts adverse outcomes independently of respiratory impairment \cite{ref15,maltais2014limb}.

Among skeletal-muscle phenotypes, muscle mass, force, and muscle quality are related but not interchangeable. Strength deficits can appear before overt atrophy and may reflect oxidative stress, mitochondrial dysfunction, impaired calcium handling, or excitation-contraction uncoupling rather than reduced tissue bulk alone \cite{ref25,ref27,ref52}. These mechanisms motivate biomarker-guided prediction of muscle outcomes, but translational biomarker studies often have small sample sizes. In such settings, unstable feature selection and data leakage can overwhelm the biological signal, so modelling claims require strong classical baselines, train/test-separated preprocessing, repeated evaluation, and conservative statistical interpretation \cite{ref59,ref72}.

\begin{figure}[!ht]
  \centering
  \includegraphics[width=0.78\textwidth]{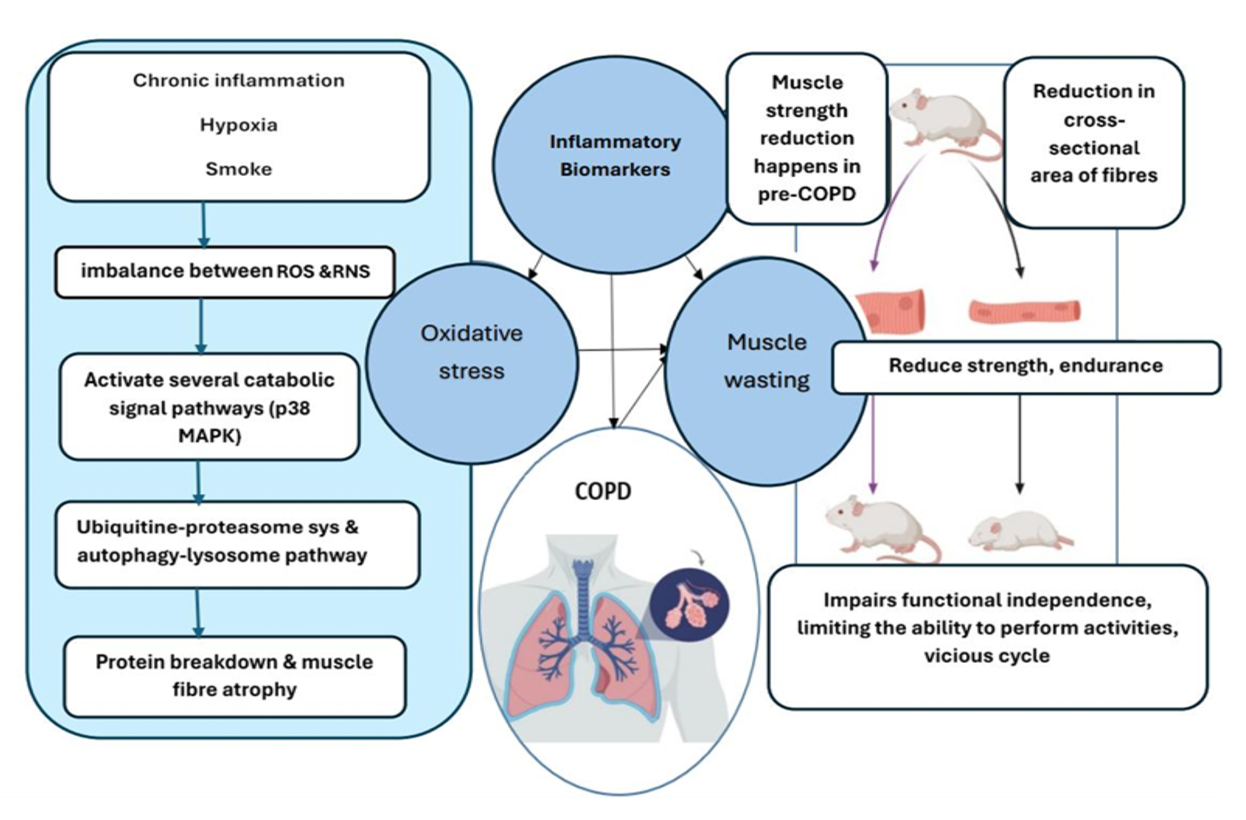}
  \caption{Mechanistic framework linking cigarette smoke exposure, inflammation, and oxidative stress to skeletal muscle dysfunction in COPD. The diagram summarises how inflammatory and redox pathways can drive catabolic signalling, muscle wasting, weakness, and functional decline. Created with BioRender.com.}
  \label{fig:copd_mechanism}
\end{figure}

\begin{revblock}
This study uses the controlled COPD animal cohort as a small-data test bed for structured nonlinear representations. We compare conventional classical models, geometry-aware symmetric positive definite (SPD) descriptors, statevector-simulated quantum-kernel ridge regression (QKR), clustered quantum-kernel features, SPD-kernel-embedded quantum-circuit models, and SPD-normalized ROSE/SteinDiv bridges. The empirical question is not whether quantum hardware provides a speed advantage; all quantum results are simulator-based. Instead, we ask whether low-dimensional SPD and quantum feature maps provide reproducible predictive structure beyond strong classical references under repeated cross-validation.

The main contributions of this study are as follows:
\begin{enumerate}
    \item We provide a condition-stratified repeated-cross-validation benchmark for three biologically distinct skeletal-muscle endpoints in experimental COPD: muscle weight, force, and muscle quality.

    \item We propose a hybrid geometry-aware quantum-kernel framework in which SPD-derived biomarker descriptors are first mapped into a reproducing-kernel Hilbert space and then normalised in that feature space before compact quantum regression. This ordering is motivated by prior work showing that quantum-kernel and variational quantum models are sensitive to the encoded feature map, input scaling, and bandwidth, which affect the induced kernel spectrum, expressivity, and generalisation behaviour \cite{schuld2021encoding,shaydulin2022bandwidth,canatar2023bandwidth}. The framework combines training-data-based synthetic SPD references, kernel-induced feature mapping, feature-space normalisation, dimensionality reduction, and low-dimensional quantum regression.

    \item We introduce a synthetic-anchor representation strategy in which real samples are represented through Stein-divergence or Stein-kernel similarities to fold-specific synthetic SPD references, reducing reliance on mixed real--synthetic landmark pools and directly testing whether synthetic SPD geometry can provide useful predictive structure.

    \item We compare direct compact quantum-kernel regression, clustered/Nyström quantum-kernel features, variational quantum readouts, SPD-kernel-embedded quantum models, and classical ridge/kernel baselines within the same repeated-cross-validation protocol, allowing the contribution of each representation family to be assessed endpoint by endpoint.

    \item We include real-to-synthetic sensitivity analyses and SPD embedding-dimension analyses to evaluate whether the observed behaviour is robust to reference construction, Hilbert-space compression, and compact quantum feature design.

    \item We provide fold-level paired statistical comparisons between the best quantum/SPD and best classical models where matched outputs are available, including Wilcoxon signed-rank testing and Holm adjustment, rather than relying only on mean cross-validation scores.

    \item We provide a biological interpretation that separates muscle force, muscle mass, and muscle quality as related but non-interchangeable COPD muscle phenotypes, showing that force is best captured by a linear biomarker model in this cohort, whereas the proposed kernel-geometric quantum representation had the numerically lowest mean RMSE for muscle weight and quality.
\end{enumerate}
\end{revblock}

\section{Background and Related Work}\label{sec:background}

\subsection{COPD as a systemic disease and the motivation for earlier risk stratification}
COPD is traditionally defined and staged using spirometry, yet clinical experience and large-scale studies consistently show that lung function alone does not fully capture symptom severity, exacerbation risk, or systemic impact \cite{ref1,ref8}. Exacerbations represent a major clinical inflection point. They are associated with worsened health status, increased healthcare utilisation, and accelerated decline, and they are commonly accompanied by a systemic inflammatory response \cite{ref4,ref5,ref55}. Biomarkers such as CRP are frequently elevated in acute exacerbation and have also been linked to worse outcomes and increased disease burden \cite{ref6,ref55}.

A growing body of evidence supports COPD as a systemic inflammatory syndrome in which pulmonary immune activation spills over into the circulation and contributes to comorbid disease \cite{ref13,ref11}. Comorbidity prevalence in COPD is high; many individuals have at least one additional chronic condition, and multi-morbidity is common \cite{ref10,ref12}. Systemic consequences are partly driven by shared risk factors such as smoking and ageing, but also by chronic inflammatory and metabolic dysregulation that persists beyond the lung \cite{ref24}. From a translational perspective, this systemic viewpoint motivates two practical needs: (i) biomarkers that reflect extrapulmonary impact and (ii) modelling approaches that integrate multiple measurements to support earlier risk stratification.

\subsection{Skeletal muscle dysfunction in COPD: mass, strength, and muscle quality}
Skeletal muscle dysfunction in COPD encompasses both quantitative loss of muscle mass (wasting) and qualitative loss of muscle function (weakness). Clinically, these manifestations contribute to reduced mobility, exercise intolerance, and impaired quality of life \cite{ref15,maltais2014limb}. Importantly, muscle loss phenotypes have been associated with adverse outcomes in large cohorts, reinforcing muscle health as a key determinant of prognosis \cite{ref17,ref21}.

A key concept for both biology and modelling is that muscle mass and strength, while related, can dissociate. Strength deficits may occur early, even before overt atrophy, and they may reflect impairments in contractile machinery, neuromuscular activation, mitochondrial function, and calcium handling rather than reductions in bulk alone \cite{ref25,ref26,ref27}. This distinction is clinically relevant because interventions that increase muscle size do not necessarily restore function. For example, pharmacological blockade of activin type II receptors can increase muscle volume in COPD, yet functional improvements may remain modest or absent \cite{ref40}. For this reason, considering muscle quality---the amount of force generated per unit muscle mass---provides an additional lens on disease impact and on potential mechanisms.

\subsection{Inflammation and oxidative stress as mechanistic drivers of muscle impairment}
COPD-associated muscle dysfunction is multifactorial, but chronic inflammation and oxidative stress repeatedly emerge as central drivers. Cigarette smoke activates airway epithelial and immune pathways, recruiting innate immune cells such as neutrophils and macrophages and sustaining cytokine production \cite{ref11,ref79}. These local inflammatory processes can propagate systemically, with cytokines and acute-phase responses contributing to a pro-catabolic peripheral environment \cite{ref13,ref29}.

At the muscle level, inflammatory mediators can activate proteolytic systems such as the ubiquitin--proteasome and autophagy--lysosome pathways, leading to net protein breakdown and fibre atrophy \cite{ref22,ref51}. Oxidative stress amplifies this process. Reactive oxygen species (ROS) and reactive nitrogen species can damage lipids, proteins, and DNA, impair mitochondrial function, and disrupt excitation--contraction coupling \cite{ref49,ref50,ref52}. Experimental and clinical evidence suggests that ROS-related signalling is a ``double-edged sword'': physiological ROS are required for normal adaptation, but excessive and sustained oxidative stress drives dysfunction \cite{ref48,ref52}. In preclinical smoke-exposure models, targeting specific sources of oxidative stress (e.g., NADPH oxidase inhibition with apocynin) can prevent loss of muscle mass and function, supporting oxidative stress as a mechanistically relevant and potentially treatable pathway \cite{ref53}.

These biological relationships motivate the biomarker space explored in this work. Markers of airway inflammation (e.g., bronchoalveolar lavage fluid cell counts), systemic inflammation (e.g., CRP), lung cytokine signalling (e.g., TNF-$\alpha$ mRNA), and muscle oxidative stress capture different layers of the inflammatory cascade and provide plausible predictors of muscle outcomes \cite{ref29,ref32,ref78}.

\subsection{Biomarker discovery: from candidate markers to multivariate signatures}
Historically, many studies have focused on individual candidate biomarkers such as TNF-$\alpha$, IL-6, IL-8, and myostatin. While biologically compelling, findings are often heterogeneous across cohorts and disease stages, and no single biomarker has achieved broad validation for COPD-associated muscle dysfunction \cite{ref22,ref29,ref33,ref77}. This limitation is not unique to muscle outcomes. COPD itself is highly heterogeneous, and biomarker expression is influenced by comorbidities, medications, and acute events \cite{ref55}.

Proteomic and multi-omics profiling has expanded the biomarker search space and has enabled the discovery of molecular subtypes and panels of circulating markers in stable COPD \cite{ref54,ref56}. Systemic proteomic signatures have also been linked to exacerbation phenotypes, highlighting the value of integrative molecular profiling \cite{ref57}. However, high-dimensional omics studies require careful computational design to avoid overfitting, and they raise a practical question for translation: how can we build robust predictors when only a modest number of variables are available in a typical clinical or experimental setting?

Preclinical models provide an important bridge. They allow controlled exposures (e.g., cigarette smoke versus air), controlled timing, and direct tissue sampling, enabling mechanistic exploration of biomarker--outcome links that are difficult to isolate in heterogeneous human cohorts \cite{ref71,ref53}. In this setting, multivariate modelling can quantify the joint predictive value of inflammatory and systemic features, identify compact signatures, and provide interpretable hypotheses for downstream mechanistic work.

\begin{revblock}
Because the present dataset is derived from a controlled animal model, the goal is not to infer a patient-ready COPD risk model. Instead, the goal is to test whether structured nonlinear representations can recover biomarker--muscle relationships under experimentally controlled exposure conditions. This distinction is important because predictors, endpoints, and sources of heterogeneity differ substantially between mouse models and human COPD cohorts.
\end{revblock}

\subsection{Machine learning for COPD biomarker modelling}
ML has become a central tool for integrating heterogeneous biomedical variables and identifying predictive signatures. In COPD, supervised models have been used for early diagnosis using quantitative imaging features \cite{ref58}, for subtype classification using multi-omics representations \cite{ref63}, and for exacerbation prediction using explainable models that quantify feature contributions \cite{ref62}. These studies highlight two practical lessons: (i) multimodal information can improve performance and (ii) interpretability is essential if the goal is biological insight rather than purely predictive accuracy.

For preclinical datasets, the main technical constraints are small sample size and strong group structure. Small $n$ increases the variance of performance estimates and can produce unstable feature selection when hyperparameters are tuned aggressively \cite{ref59,ref72}. Group structure (e.g., Sham vs cigarette smoke exposure) can dominate the signal and may lead to models that implicitly ``memorise'' condition rather than learn biomarker relationships. These challenges make evaluation design a first-order concern.

\begin{revblock}
In addition to standard regression metrics, many translational questions are screening-oriented, but the present screening view is defined within the experimental cohort rather than by a validated human clinical cutoff. If a biologically defined threshold in the experimental cohort defines ``low'' outcomes, continuous predictions can be evaluated for their ability to rank low-outcome animals. Importantly, the threshold itself must be derived from the training split only; otherwise, screening metrics become optimistically biased \cite{ref7}. This paper adopts that principle explicitly.
\end{revblock}

\subsection{Geometry-aware representations with symmetric positive definite matrices}
Beyond standard vectorial representations, a growing body of work encodes observations using symmetric positive definite (SPD) matrices, such as covariance descriptors, and exploits the geometry of the SPD manifold \cite{ref83}. SPD matrices do not form a Euclidean space under ordinary arithmetic, and naive distances can distort structure. Geometry-aware approaches either embed SPD matrices into tangent spaces or use kernel-based comparisons that respect manifold properties \cite{ref83}.

\begin{revblock}
A practical and widely used geometry-aware dissimilarity measure on SPD matrices is the Jensen--Bregman LogDet divergence, also known as the Stein divergence \cite{sra2012newmetric}. For SPD matrices $\mathbf{A}$ and $\mathbf{B}$, it is defined as
\begin{equation}
\label{eq:stein_bg}
 d_{\mathrm{Stein}}(\mathbf{A},\mathbf{B})
= \log\det\!\Big(\tfrac{\mathbf{A}+\mathbf{B}}{2}\Big)
- \tfrac{1}{2}\log\det(\mathbf{A})
- \tfrac{1}{2}\log\det(\mathbf{B}),
\end{equation}
which is symmetric and non-negative. In small biomedical datasets, an appealing strategy is to construct an SPD descriptor per sample that captures second-order structure among biomarkers (for example, outer products of normalised biomarker vectors, or local covariance estimates from neighbourhoods in feature space). A low-dimensional embedding can then be formed by measuring Stein divergences to a small set of representative prototypes.

To obtain prototypes without relying on Euclidean structure, we use $K$-medoids clustering (PAM) on a pairwise divergence matrix \cite{kaufman1990finding}. $K$-medoids selects representatives from the supplied SPD pool. When the pool contains only real training descriptors, the medoids are observed training SPD matrices; when synthetic augmentation is enabled, selected medoids may be either real training descriptors or training-derived synthetic SPD descriptors. Each sample is then represented by its vector of Stein-divergence features relative to the medoids, yielding a compact feature representation for standard downstream regressors.

When data are scarce, estimating stable prototypes can be difficult. A practical regularisation is to expand the clustering pool with synthetic SPD matrices generated by interpolation between training descriptors in a log-Euclidean domain \cite{arsigny2006logeuclidean}. Log-Euclidean interpolation preserves positive definiteness and produces plausible intermediate descriptors while all synthetic matrices are generated from the training split only. This combination---Stein divergence, $K$-medoids prototypes, and log-Euclidean augmentation---provides a geometry-aware representation that is computationally compact, interpretable, and well matched to low-$n$ settings.
\end{revblock}

\subsection{Quantum kernels and variational quantum regression for low-dimensional tabular data}
Quantum machine learning provides a principled way to construct feature maps and kernel functions using quantum states \cite{ref88}. In a quantum-kernel approach, a classical input $\mathbf{x}$ is encoded into a quantum state $|\phi(\mathbf{x})\rangle$ through a feature-map circuit. Similarities are computed via state fidelity,
\begin{equation}
\label{eq:fidelity}
 k(\mathbf{x},\mathbf{z}) = \big|\langle \phi(\mathbf{x})\,|\,\phi(\mathbf{z}) \rangle\big|^2,
\end{equation}
which defines a valid kernel for kernel ridge regression under appropriate regularisation \cite{Havlicek2019,SchuldKilloran2019}. Conceptually, the feature map induces a high-dimensional (often implicit) feature space in which a linear model is fit. This can be attractive for small-sample settings because the non-linearity is expressed through the kernel rather than through many trainable parameters.

In practice, quantum kernels can become overly ``peaky'' (near-identity), meaning that test samples may have near-zero similarity to training samples after encoding. This behaviour can make the kernel matrix ill-conditioned and can lead to unstable regression. One way to stabilise similarity-based learning is to replace the full $n\times n$ kernel representation with a compact set of similarities to representative centres. This idea is closely related to the Nystr\"om approximation for kernel machines \cite{WilliamsSeeger2001}.

Motivated by this, we consider a clustered quantum-kernel-feature construction. A small set of centres is learned from the training split (e.g., via k-means in a low-dimensional parameter space), and each sample is represented as a $k$-dimensional vector $[k(\mathbf{x},\mathbf{c}_1),\ldots,k(\mathbf{x},\mathbf{c}_k)]$. An optional whitening step based on the centre--centre kernel can further stabilise feature geometry and reduce the impact of kernel concentration. This produces a compact, regularised representation that can be used with simple ridge regression.

In this work, the primary empirical comparison focuses on kernel-based quantum models. This keeps the number of trainable parameters low and focuses the analysis on whether structured quantum feature maps provide useful nonlinear similarity measures in the present small-sample setting. We also retain a compact variational quantum regressor as a supplementary analysis for the trainable-circuit branch. Because VQR introduces additional stochastic optimisation and circuit-training choices, it is not used for the primary inferential ranking, but it is reported to keep the quantum modelling scope transparent.

\section{Materials and Methods}

\subsection{Problem formulation and notation}
Let $n$ denote the number of experimental subjects (animals), and let
$\{(\mathbf{x}_i, c_i, y_i)\}_{i=1}^{n}$ denote the dataset, where:
(i) $\mathbf{x}_i \in \mathbb{R}^{d}$ is the vector of measured biomarkers and physiological covariates,
(ii) $c_i \in \{0,1\}$ encodes the experimental condition (e.g., Sham vs.\ chronic smoke exposure),
and (iii) $y_i \in \mathbb{R}$ is a continuous target variable.

We considered three regression targets:
\begin{enumerate}
    \item \textbf{Muscle mass:} Tibialis Anterior (TA) muscle weight, $y^{(\mathrm{w})}$ (mg),
    \item \textbf{Muscle function readout:} tibialis anterior force readout, $y^{(\mathrm{f})}$ (mN; referred to as specific force in the source dataset),
    \item \textbf{Muscle quality index:} $y^{(\mathrm{q})} = y^{(\mathrm{f})} / y^{(\mathrm{w})}$ (mN per mg).
\end{enumerate}
The quality index is computed per subject using the  measured force and weight.

All learning problems are posed as supervised regression. In addition, we report a screening-style Receiver-Operating Characteristic Area Under the Curve (ROC-AUC) by thresholding the continuous target into a binary label (``low'' vs.\ ``not-low''); this is described in Section~\ref{subsec:metrics}.

\subsection{Dataset and predictors}
The dataset contains $n=213$ subjects with the following observed variables:
experimental condition (Sham vs.\ chronic smoke exposure), bronchoalveolar lavage fluid (BALF)
cell counts (total cells, macrophages, neutrophils, lymphocytes), systemic inflammation marker
C-reactive protein (CRP), muscle oxidative stress, lung tumor necrosis factor alpha (TNF-$\alpha$) mRNA,
and physiological measures including oxygen consumption ($\mathrm{VO}_2$) and locomotor activity.
These measurements and the biomedical context are described in detail in the associated thesis. \cite{KhaliliThesis}
Animal procedures and ethics approvals are described in the experimental work underlying this dataset; see \cite{KhaliliThesis} and references therein.

\subsection{Animal ethics and experimental oversight}
Male BALB/c mice (7 weeks of age) were obtained from the Animal Resources Centre (Perth, WA, Australia) and housed in micro-isolator cages (four mice per cage) at $21 \pm 0.5\,^{\circ}\mathrm{C}$ on a 12-h day/night cycle with ad libitum access to standard chow and water. After one week of acclimatisation, mice were randomly assigned to room-air (Sham) or cigarette-smoke exposure groups. Animals were weighed three times per week and monitored daily. All experiments were conducted in accordance with the Australian Code of Practice for the Care of Experimental Animals and with RMIT University Animal Ethics Committee approval (Animal Ethics Application Number 1928). Reporting follows the ARRIVE 2.0 guidelines and the recommendations of the British Journal of Pharmacology \cite{perciedusert2020arrive,lilley2020arrive}. The cigarette-smoke model used here is part of a broader and well-characterised preclinical COPD program in which lung inflammation, emphysematous remodelling, and related physiological impairments have been quantified in detail \cite{ref53,ref71,ruwanpura2016}.

\subsection{Train/test protocol}
\begin{revblock}
All main-text results use repeated stratified five-fold cross-validation with three repeats, giving 15 matched held-out evaluations per endpoint. Each outer split held out approximately 20\% of the animals and was stratified by experimental condition to preserve Sham/CS balance. Let $\mathcal{I}_{\mathrm{tr}}$ and $\mathcal{I}_{\mathrm{te}}$ denote the training and test indices for one outer split.
\end{revblock}

\begin{revblock}
For every outer split, all data-dependent operations were fit from $\mathcal{I}_{\mathrm{tr}}$ only and then applied once to $\mathcal{I}_{\mathrm{te}}$. This included imputation, Yeo--Johnson transformation, scaling, PCA or angle scaling, SPD descriptor construction, synthetic SPD generation, prototype or reference selection, kernel centering statistics, threshold definitions for screening metrics, and all hyperparameter selection. Inner cross-validation used only the corresponding outer-training partition. No held-out target values, held-out covariates, or held-out SPD descriptors were used to generate synthetic matrices, construct reference sets, fit Hilbert-space normalizers, choose hyperparameters, or define screening thresholds.
\end{revblock}

\begin{revblock}
We denote the resulting inner-training and inner-validation indices by $\mathcal{I}^{(k)}_{\mathrm{tr}}$ and $\mathcal{I}^{(k)}_{\mathrm{va}}$, respectively. Reported tables average fold-level metrics across the 15 held-out evaluations.
\end{revblock}

\subsection{Feature preprocessing}
\label{sec:feature_preprocessing}
\begin{revblock}
\subsubsection{Condition encoding}
Continuous predictors were processed using training-fold imputation, Yeo--Johnson transformation, and scaling. The binary condition indicator was encoded as a numeric covariate when included. For the SPD/RKHS bridge models, condition was appended after the geometry-induced representation was constructed; for compact vector and kernel baselines, condition was included in the train-only numeric preprocessing pipeline. Condition-excluded variants omitted $c_i$.

Equivalently, compact vector/kernel baselines used a train-fitted numeric preprocessing map of the form
\[
\mathbf{z}_i = \Phi_{\mathrm{vec}}\left([\mathbf{x}_i^\top,\; c_i]^\top\right),
\]
whereas the SPD/RKHS bridge used
\[
\tilde{\mathbf{z}}_i = [g_{\mathrm{SPD}}(\Phi_x(\mathbf{x}_i))^\top,\; c_i]^\top,
\]
with $g_{\mathrm{SPD}}$ denoting the train-only geometry-induced representation. In all cases, preprocessing maps, geometry operators, and scaling parameters were fitted only on the corresponding training partition.

\subsubsection{Power transform and scaling}
To mitigate skewness and heteroscedasticity in biomedical measurements, we applied the \textbf{Yeo--Johnson power transform} feature-wise within each train-only numeric preprocessing pipeline \cite{yeojohnson2000}. Let $T_{\mathrm{YJ}}(\cdot;\lambda)$ denote the Yeo--Johnson transformation with parameter $\lambda$ estimated from training data by maximum likelihood. After power transformation, we used either: (i) standard scaling (zero mean, unit variance) or (ii) robust scaling (median/IQR scaling), both fit on training data only. In compact vector/kernel baselines, the encoded condition indicator was treated as part of the numeric feature matrix when condition was included; in the SPD/RKHS bridge, condition was not used to construct the SPD descriptors or synthetic/RKHS references and was appended after the geometry-induced coordinates were formed.
\end{revblock}

\begin{revblock}
\subsubsection{Target transform (log1p)}
When enabled, the regression target was transformed as
\[
\tilde{y}_i = \log(1+y_i).
\]
Models were trained to predict $\tilde{y}_i$. Predictions were then transformed back to physical units by
\[
\hat{y}_i = \exp(\widehat{\tilde{y}}_i)-1.
\]
All RMSE, MAE, $R^2$, percent RMSE, and ROC--AUC values reported in the tables were computed using predictions in physical target units.
\end{revblock}

\subsubsection{Dimensionality reduction for quantum models}
To map features to a small number of qubits, we optionally used \textbf{principal component analysis (PCA)}
fit on training data, selecting $q \le p$ principal components. The reduced representation is:
\[
\mathbf{u}_i \in \mathbb{R}^{q}.
\]
We then rescaled each component to a bounded angle interval
(e.g.\ $[-\pi/2,\pi/2]$) using a min--max map computed on training data:
\[
\boldsymbol{\theta}_i \in [\theta_{\min}, \theta_{\max}]^q.
\]
These angles are the inputs to quantum circuits.

\subsection{Classical baseline models}
\label{sec:blockA}

\subsubsection{Overview}
We evaluate classical baseline predictors and regression models for the three targets under the shared outer-fold protocol and preprocessing described above. In addition to global-mean and condition-means baselines, we evaluate ridge regression, random forest regression, a shallow decision tree, and a simple condition-axis baseline that regresses on the one-dimensional LDA discriminant score separating Sham vs.\ CS (``LDA condition axis then Ridge'').

\subsubsection{Baseline predictors}
We report two simple reference predictors:
\begin{enumerate}
\item Global-mean baseline: $\hat{y}=\frac{1}{|\mathcal{T}|}\sum_{i\in \mathcal{T}} y_i$, where $\mathcal{T}$ is the training set.
\item Condition-means baseline: $\hat{y}=\mu_{\text{Sham}}$ if $c=\text{Sham}$ and $\hat{y}=\mu_{\text{CS}}$ if $c=\text{CS}$, where $\mu_{\text{Sham}}$ and $\mu_{\text{CS}}$ are computed on the training set.
\end{enumerate}

\subsubsection{Classical regression models}
We evaluate three standard regression models.

\paragraph{Ridge regression.}
Ridge regression \cite{hoerl1970ridge} fits a linear predictor $\hat{y}=\mathbf{w}^{\top}\mathbf{x}+b$ by minimizing
\begin{equation}
\min_{\mathbf{w},b}\sum_{i\in \mathcal{T}} \left(y_i^{(\mathrm{tr})} - (\mathbf{w}^{\top}\tilde{\mathbf{x}}_i+b)\right)^2
+ \alpha \|\mathbf{w}\|_2^2,
\end{equation}
where $\alpha>0$ controls $\ell_2$ regularization strength.

\paragraph{Random forest regression.}
Random forests \cite{Breiman2001} combine $T$ regression trees trained on bootstrap samples with randomized feature selection at splits. The prediction is the ensemble average of tree predictions.

\paragraph{Shallow decision tree regression.}
We also train an interpretable CART-style regression tree \cite{breiman1984cart}, restricted to small depth and minimum leaf sizes to improve interpretability and reduce variance.

\subsubsection{Hyperparameter selection by cross-validation}
For each model family, the candidate grid was evaluated within the corresponding outer-training partition and the selected configuration was then assessed on the held-out outer fold.
The selection criterion is root mean squared error (RMSE) measured in the physical target units by applying the inverse response transform.
 After hyperparameter selection, the chosen model is refit on the full  outer-training set and evaluated on the held-out outer fold test set.

\subsubsection{Evaluation metrics}
\begin{revblock}
We report RMSE and $R^2$ for each outer held-out test fold, and then average these quantities across the 15 repeated cross validation evaluations (consist of 3 times repeated 5 fold cross validation); formal definitions are given in Section~\ref{subsec:metrics}. For the auxiliary screening view, we report ROC-AUC using the thresholding protocol described in Section~\ref{subsec:metrics}, with threshold $\tau=0.8\,\mathrm{mean}(y\mid\mathrm{Sham})$ and positive class ``low''.
\end{revblock}

\subsection{Classical feature engineering and condition interactions}
\label{sec:blockB}

\subsubsection{Motivation}
These results show that classical models already perform competitively on muscle weight and force readout, with the experimental condition contributing substantial explanatory power.
Next, we test whether a conservative, mechanistically motivated feature expansion can improve predictive performance under the same train/test protocol.
The goal is not to increase model capacity arbitrarily, but to introduce low-dimensional composite covariates that reflect well-known inflammatory and physiological couplings (ratios and bilinear products), and to allow feature effects to differ between Sham and CS via interaction terms.

\subsubsection{Engineered feature map}
Let $\mathbf{x}_i\in\mathbb{R}^{d}$ denote the raw feature vector for subject $i$ after median imputation of missing values (computed from the training set only).
We define an engineered feature map $\phi:\mathbb{R}^{d}\rightarrow\mathbb{R}^{d'}$ by augmenting $\mathbf{x}_i$ with the following composite covariates (when the required variables are present in the dataset).

Let $N_i$ be neutrophils count, $L_i$ lymphocytes count, $\mathrm{CRP}_i$ blood C-reactive protein, $B_i$ total bronchoalveolar lavage fluid (BALF) cell count, $S_i$ muscle oxidative stress, $V_i$ oxygen consumption $\mathrm{VO}_2$, and $T_i$ lung tumor necrosis factor alpha messenger ribonucleic acid (TNF$\alpha$ mRNA) fold-change from Sham.
Using a small constant $\varepsilon=10^{-9}$ to prevent division by zero, we add:
\begin{align}
\mathrm{NLR}_i &= \frac{N_i}{L_i+\varepsilon},\\
\mathrm{CRPperCell}_i &= \frac{\mathrm{CRP}_i}{B_i+\varepsilon},\\
\mathrm{OxStressOverVO2}_i &= \frac{S_i}{V_i+\varepsilon},\\
\mathrm{CRPVO2}_i &= \mathrm{CRP}_i \cdot V_i,\\
\mathrm{CRPOxStress}_i &= \mathrm{CRP}_i \cdot S_i,\\
\mathrm{TNFaNeutrophils}_i &= T_i \cdot N_i.
\end{align}
All engineered features are deterministic functions of the subject’s covariates and do not use label information.

\subsubsection{Condition interactions}
Let $c_i\in\{0,1\}$ denote the condition indicator, where $c_i=0$ for Sham and $c_i=1$ for CS.
We include $c_i$ as an additional predictor and also introduce interaction features:
\begin{equation}
\phi_{\times c}(\mathbf{x}_i,c_i) = \big[\phi(\mathbf{x}_i),\; c_i,\; c_i\cdot \phi(\mathbf{x}_i)\big].
\end{equation}
This expansion allows the model to represent condition-specific linear effects by learning separate slopes for CS relative to Sham.

\subsubsection{Preprocessing, model selection, and evaluation}
\begin{revblock}
All subsequent steps follow the classical baseline protocol exactly:
(i) the engineered numeric feature matrix, including encoded condition and interaction terms when present, is transformed using train-only Yeo--Johnson preprocessing and scaling;
(ii) transformed inputs are standardized using training-set mean and standard deviation;
(iii) the regression target is trained in $\log(1+y)$ space and inverted back to physical units for reporting;
(iv) hyperparameters are selected using the train-only inner-cross-validation protocol within each outer fold's training set; and
(v) reported metrics are computed by averaging performance across outer held-out folds. For evaluation, a threshold $\tau=0.8\,\mathrm{mean}(y\mid\mathrm{Sham})$ is computed from the training partition, the positive class is defined as ``low'', and ROC-AUC is derived from the continuous predictions.
\end{revblock}

\subsection{Geometry-informed mapping on the manifold of symmetric positive definite matrices}
\label{sec:blockC_spd}

\subsubsection{Motivation and overview}
\begin{revblock}
For small tabular datasets, second-order interactions between covariates can be informative, but explicitly enumerating interaction terms can rapidly increase dimensionality and overfitting risk. To introduce second-order structure while preserving a controlled model capacity, we construct a symmetric positive definite (SPD) matrix descriptor for each sample and exploit the geometry of the SPD manifold. The core idea is to map each sample to a low-dimensional vector of Stein-divergence features relative to representative SPD prototypes (cluster centres), then concatenate this divergence-feature vector with the input covariates and train a regularised linear regressor.
\end{revblock}

Let $i \in \{1,\ldots,n\}$ index samples. After train-only preprocessing (Section~\ref{sec:feature_preprocessing}), each sample is represented by a feature vector
\begin{equation}
\mathbf{x}_i \in \mathbb{R}^{p},
\end{equation}
where $p$ denotes the dimension of the compact representation used by the relevant model branch. In the COPD experiments reported here, we used a fixed compact covariate subset to keep the geometric representation low-dimensional and numerically stable.
\begin{revblock}
For standalone SPD-distance benchmark rows, condition-enabled variants used three selected continuous biomarkers together with the encoded condition, yielding a compact $4\times4$ outer-product SPD descriptor; condition-excluded variants used the corresponding $3\times3$ biomarker descriptor. For the SPD/RKHS bridge models, including Synth\_ROSE and SteinDiv-Hilbert-QKR, SPD descriptors and synthetic/RKHS references were constructed from the continuous biomarker representation, and the encoded condition indicator was appended only after the geometry-induced coordinates had been formed. This intentionally small descriptor size limits computational overhead, and the SPD branch serves as a transparent geometry-aware benchmark rather than a requirement for more complex geometric modelling.

The compact continuous biomarker panel was endpoint-specific: muscle weight used C-reactive protein, neutrophil count, and total bronchoalveolar-lavage fluid cellularity; force used muscle oxidative stress, lung TNF-\(\alpha\) mRNA, and neutrophil count; and muscle quality used lung TNF-\(\alpha\) mRNA, neutrophil count, and lymphocyte count. These compact panels were retained as fixed biologically interpretable design choices within each endpoint and reused unchanged across outer folds rather than re-selected within the outer cross-validation loop. The encoded condition indicator was included according to the branch-specific preprocessing policy described in Section~\ref{sec:feature_preprocessing}. The associated feature-set sensitivity audit is retained in the supplementary audit materials rather than used for main-text model ranking.
\end{revblock}

\subsubsection{SPD descriptors}
An SPD matrix is a symmetric matrix $\mathbf{S}\in\mathbb{R}^{p\times p}$ satisfying $\mathbf{v}^{\top}\mathbf{S}\mathbf{v}>0$ for all nonzero $\mathbf{v}\in\mathbb{R}^{p}$ \cite{bhatia2007positive}. We consider two SPD constructions.

\paragraph{(i) Outer-product SPD descriptor (interaction descriptor).}
Given $\mathbf{x}_i$, we optionally normalise to unit Euclidean norm,
\begin{equation}
\tilde{\mathbf{x}}_i = \frac{\mathbf{x}_i}{\|\mathbf{x}_i\|_2 + \delta},
\end{equation}
with a small $\delta>0$ to avoid division by zero. The SPD descriptor is then defined as
\begin{equation}
\mathbf{S}_i = \tilde{\mathbf{x}}_i \tilde{\mathbf{x}}_i^{\top} + \varepsilon \mathbf{I}_{p},
\label{eq:outer_spd}
\end{equation}
where $\varepsilon>0$ is a diagonal jitter and $\mathbf{I}_{p}$ is the $p\times p$ identity matrix. The rank-one term $\tilde{\mathbf{x}}_i \tilde{\mathbf{x}}_i^{\top}$ encodes second-order interactions between components of $\mathbf{x}_i$, while $\varepsilon\mathbf{I}_{p}$ ensures strict positive definiteness.
\begin{revblock}
Because the outer-product descriptor is rank one before adding the diagonal jitter, it is intentionally near-rank-deficient; for the compact $4\times4$ descriptor, three eigenvalues are governed mainly by the jitter floor. Stein-divergence values can therefore depend partly on the selected $\varepsilon$, which is why the local-neighbourhood covariance construction in Equation~\eqref{eq:localcov_spd} is retained as a full-rank SPD variant for sensitivity and methodological comparison.
\end{revblock}

\paragraph{(ii) Local-neighbourhood covariance SPD descriptor.}
To approximate a ``local covariance'' structure, for each training sample $i$ we form a neighbourhood $\mathcal{N}_i$ consisting of the $k_{\mathrm{NN}}$ nearest neighbours of $\mathbf{x}_i$ under Euclidean distance in the transformed feature space. Let $\bar{\mathbf{x}}_i$ denote the neighbourhood mean and define the empirical covariance
\begin{equation}
\mathbf{\Sigma}_i
=
\frac{1}{|\mathcal{N}_i|-1}\sum_{j\in\mathcal{N}_i}
(\mathbf{x}_j-\bar{\mathbf{x}}_i)(\mathbf{x}_j-\bar{\mathbf{x}}_i)^{\top}.
\end{equation}
To improve conditioning in small samples, we apply shrinkage toward a scaled identity (Ledoit--Wolf style) \cite{ledoit2004well}
\begin{equation}
\mathbf{S}_i
=
(1-\lambda)\mathbf{\Sigma}_i
+
\lambda\frac{\mathrm{tr}(\mathbf{\Sigma}_i)}{p}\mathbf{I}_{p}
+
\varepsilon \mathbf{I}_{p},
\label{eq:localcov_spd}
\end{equation}
with $\lambda\in[0,1]$ and $\varepsilon>0$. For test samples, neighbourhoods are formed with respect to the training set only to preserve a strict train/test separation.

\subsubsection{Stein divergence on SPD matrices}
To compare SPD matrices we use the symmetric Stein divergence (also known as the Jensen--Bregman LogDet divergence) \cite{sra2012newmetric,cherian2011jblD}:
\begin{equation}
D_{\mathrm{Stein}}(\mathbf{A},\mathbf{B})
=
\log\det\left(\frac{\mathbf{A}+\mathbf{B}}{2}\right)
-
\frac{1}{2}\log\det(\mathbf{A}\mathbf{B}),
\label{eq:stein}
\end{equation}
\begin{revblock}
defined for SPD matrices $\mathbf{A},\mathbf{B} \in \mathbb{S}_{++}^{p}$. This divergence is symmetric, nonnegative, and empirically effective for learning tasks on SPD manifolds. In our implementation, positive definiteness is guaranteed by the diagonal jitter $\varepsilon\mathbf{I}_p$ in Equations~\eqref{eq:outer_spd}--\eqref{eq:localcov_spd}. This form is algebraically equivalent to Equation~\eqref{eq:stein_bg}, because for SPD matrices $\log\det(\mathbf{A}\mathbf{B})=\log\det(\mathbf{A})+\log\det(\mathbf{B})$.
\end{revblock}

\subsubsection{Synthetic SPD augmentation for stable prototype selection}
To reduce prototype instability when $n$ is small, we optionally augment the training SPD set with unlabeled synthetic SPD matrices. Synthetic SPD matrices are used \emph{only} to stabilise clustering and are never assigned target labels. In the reported experiments, synthetic SPD samples are generated by geodesic interpolation in the Log-Euclidean geometry \cite{arsigny2006logeuclidean}, which preserves positive definiteness:
\begin{equation}
\mathbf{S}_{\mathrm{syn}}
=
\exp\Big((1-t)\log(\mathbf{S}_a) + t\log(\mathbf{S}_b)\Big),
\quad t\sim \mathrm{Uniform}(0,1),
\label{eq:loge_interp}
\end{equation}
where $\mathbf{S}_a$ and $\mathbf{S}_b$ are randomly selected training SPD matrices, and $\log(\cdot)$ and $\exp(\cdot)$ denote the matrix logarithm and exponential. This strategy is consistent with prior work on SPD manifold learning and random projection methods on SPD spaces \cite{ref83}.

Let $\mathcal{S}_{\mathrm{train}}=\{\mathbf{S}_i\}_{i\in\mathrm{train}}$ be the training SPD set and $\mathcal{S}_{\mathrm{syn}}$ be the synthetic SPD set. The clustering pool is
\begin{equation}
\mathcal{S}_{\mathrm{pool}} = \mathcal{S}_{\mathrm{train}} \cup \mathcal{S}_{\mathrm{syn}}.
\end{equation}

\begin{revblock}
\subsubsection{Clustering on the SPD space and Stein-divergence feature mapping}
Given $\mathcal{S}_{\mathrm{pool}}$, we compute the pairwise Stein divergence matrix using Equation~\eqref{eq:stein}, then select $K$ representative prototypes via $K$-medoids clustering (Partitioning Around Medoids) \cite{kaufman1990finding}. Unlike $K$-means, $K$-medoids returns centres that are valid SPD matrices drawn from $\mathcal{S}_{\mathrm{pool}}$; therefore, when synthetic augmentation is enabled, selected medoids may be either real training descriptors or training-derived synthetic SPD descriptors. Let $\{\mathbf{C}_1,\ldots,\mathbf{C}_K\}$ denote the selected medoids.

Each sample is mapped to a $K$-dimensional Stein-divergence feature vector
\begin{equation}
\mathbf{d}_i
=
\big[
D_{\mathrm{Stein}}(\mathbf{S}_i,\mathbf{C}_1),
\ldots,
D_{\mathrm{Stein}}(\mathbf{S}_i,\mathbf{C}_K)
\big]^{\top}
\in\mathbb{R}^{K}.
\label{eq:stein_divergence_feature_vector}
\end{equation}
Crucially, $\{\mathbf{C}_k\}_{k=1}^{K}$ are computed using training data (and optional synthetic augmentation) only. Test samples are mapped by Equation~\eqref{eq:stein_divergence_feature_vector} using the fixed training prototypes, ensuring that prototypes are learned from the training split.
\end{revblock}

\begin{revblock}
\subsubsection{Regression model with SPD divergence features}
Finally, we form the augmented representation
\begin{equation}
\mathbf{h}_i = [\mathbf{x}_i^{\top},\mathbf{d}_i^{\top}]^{\top}\in\mathbb{R}^{p+K},
\label{eq:augmented_feature}
\end{equation}
and fit ridge regression
\begin{equation}
\hat{y}_i = \mathbf{w}^{\top}\mathbf{h}_i + b,
\end{equation}
by minimising
\begin{equation}
\min_{\mathbf{w},b}\;
\sum_{i\in\mathrm{train}}\big(y_i-(\mathbf{w}^{\top}\mathbf{h}_i+b)\big)^2
+
\alpha\|\mathbf{w}\|_2^2,
\label{eq:ridge}
\end{equation}
where $\alpha>0$ is selected within each outer-training set. Target transformations (e.g., $\log(1+y)$) are estimated from the training partition, and reported metrics are computed in physical units after inverse transformation and then averaged across outer folds.
\end{revblock}

\subsubsection{Evaluation metrics}
We report RMSE, $R^2$, and scale-normalised percent errors (\%RMSE ) on each held-out  outer fold and then average these quantities as defined in Section~\ref{subsec:metrics}.

\subsection{Quantum machine learning models}
\label{subsec:qml}

\subsubsection{Quantum feature map and state preparation}
Quantum encoding is represented as a nonlinear feature map into a Hilbert space
of $q$ qubits. \cite{SchuldKilloran2019,Havlicek2019}
After PCA and min--max scaling, each sample yields
$\boldsymbol{\theta}_i \in [\theta_{\min},\theta_{\max}]^q$.

We define a parameterized quantum circuit $U(\boldsymbol{\theta})$ acting on
$|0\rangle^{\otimes q}$, producing a pure state
\[
|\psi(\boldsymbol{\theta})\rangle = U(\boldsymbol{\theta}) |0\rangle^{\otimes q}.
\]
The circuit uses repeated \emph{data re-uploading layers}:
in each layer, each qubit receives rotations whose angles are proportional to
the components of $\boldsymbol{\theta}$, followed by an entangling pattern
(e.g.\ a ring of controlled-NOT gates).
A global angle-scale factor $s>0$ multiplies input angles to control circuit sensitivity.
\begin{revblock}
\paragraph{Primary circuit architecture and angle encoding.}
In the compact QKR setting, each endpoint used three endpoint-specific continuous biomarkers plus the binary condition indicator. These four numeric inputs were mapped to four qubits after train-only imputation, Yeo--Johnson transformation, standardisation, optional PCA if required by the qubit budget, and min--max angle scaling to $[-\pi/2,\pi/2]$. Thus, the compact vector/QKR setting used $q=4$ qubits and input angles $\theta_j$ derived entirely from the corresponding outer-training split. The primary quantum feature map used data-reuploading layers of the form
\[
U(\boldsymbol{\theta}) = \left[E_{\mathrm{ring}} \prod_{j=0}^{q-1} R_z^{(j)}(s\theta_j) R_y^{(j)}(s\theta_j)\right]^L,
\]
where $s$ is the global angle-scale factor and $E_{\mathrm{ring}}$ is a CNOT ring applied after the single-qubit rotations. For the four-qubit compact setting, the CNOT pattern was $0\to1$, $1\to2$, $2\to3$, and $3\to0$. The QKR kernel was the squared state fidelity in Equation~\eqref{eq:qkernel}, optionally with kernel-power rescaling before centering. Angle encoding was selected as the primary representation because it maps bounded tabular variables directly into single-qubit rotations, requires only one qubit per retained feature, avoids the normalisation and state-preparation overhead associated with amplitude encoding, and yields a shallow, transparent circuit family. Alternative encodings and entanglement patterns are evaluated in the encoding/topology sensitivity analysis.
\end{revblock}

\subsubsection{Quantum-kernel ridge regression (QKR)}
\begin{revblock}
We define the \textbf{fidelity kernel} between two samples $i,j$ as
\begin{equation}
k(\boldsymbol{\theta}_i,\boldsymbol{\theta}_j)
= \left|\langle \psi(\boldsymbol{\theta}_i) \mid \psi(\boldsymbol{\theta}_j)\rangle\right|^2,
\label{eq:qkernel}
\end{equation}
For selected QKR configurations, we applied an elementwise kernel-power rescaling,
\begin{equation}
K_{ij}^{(\rho)} = K_{ij}^{\rho}, \qquad 0<\rho\le 1,
\label{eq:kernel_power_rescaling}
\end{equation}
to reduce overly sharp fidelity kernels by increasing off-diagonal similarities. This operation was treated as a finite-sample similarity rescaling rather than as a guaranteed Mercer-kernel transformation, because fractional Hadamard powers do not preserve positive semidefiniteness for all finite positive-semidefinite matrices. After rescaling, each training Gram matrix was symmetrised and its eigenspectrum was checked numerically. When explicit coordinates were required, only non-negative eigencomponents above numerical tolerance were retained; for kernel-ridge solves, ridge regularisation was applied to the finite training matrix selected within the outer-training split.
\end{revblock}

Let $\mathbf{K}\in\mathbb{R}^{n_{\mathrm{tr}}\times n_{\mathrm{tr}}}$ be the training Gram matrix. When centering was used, centering statistics and the target mean were computed within the outer-training split:
\[
\mathbf{K}_c = \mathbf{H}\mathbf{K}\mathbf{H},\quad
\mathbf{H}=\mathbf{I}-\frac{1}{n_{\mathrm{tr}}}\mathbf{1}\mathbf{1}^\top,\quad
\mathbf{y}_c=\mathbf{y}-\bar{y}_{\mathrm{train}}\mathbf{1}.
\]
\begin{revblock}
For a held-out sample, let $\mathbf{k}_*\in\mathbb{R}^{n_{\mathrm{tr}}}$ denote the uncentered vector of similarities between the test sample and all training samples. The centered train--test vector was computed using training-fold centering statistics only:
\begin{equation}
\mathbf{k}_{c,*}
=
\mathbf{k}_*
-
\frac{1}{n_{\mathrm{tr}}}\mathbf{K}\mathbf{1}
-
\frac{1}{n_{\mathrm{tr}}}\mathbf{1}\mathbf{1}^{\top}\mathbf{k}_*
+
\frac{1}{n_{\mathrm{tr}}^2}\mathbf{1}\mathbf{1}^{\top}\mathbf{K}\mathbf{1}.
\label{eq:centered_train_test_kernel_vector}
\end{equation}
\end{revblock}
QKR then solves the kernel ridge system
\[
\boldsymbol{\alpha} = (\mathbf{K}_c + \lambda \mathbf{I})^{-1}\mathbf{y}_c,
\]
and predicts for a test sample $\boldsymbol{\theta}_*$ via the centered train-test kernel vector $\mathbf{k}_{c,*}$ as
\[
\hat{y}_* = \bar{y}_{\mathrm{train}}+\mathbf{k}_{c,*}^\top \boldsymbol{\alpha}.
\]
When a candidate setting used the uncentered kernel, the analogous uncentered ridge system was applied consistently with the same train-only regularisation selection.

\subsubsection{Clustered quantum kernel-features (QKF)}
\label{subsec:qkf}
Empirically, full QKR can overfit or become numerically unstable on small datasets.
We therefore evaluate a clustered/Nyström-style approximation that maps each input to its
kernel similarities against a small set of representative centers.

First, we run $K$-means in angle space on training angles $\{\boldsymbol{\theta}_i\}_{i\in\mathcal{I}_{\mathrm{tr}}}$
to obtain centers $\{\boldsymbol{\mu}_r\}_{r=1}^{K}$.

Next, we construct \textbf{quantum kernel-features} for each sample:
\[
\boldsymbol{\phi}(\boldsymbol{\theta})
= \left[
k(\boldsymbol{\theta},\boldsymbol{\mu}_1),\;
k(\boldsymbol{\theta},\boldsymbol{\mu}_2),\;\ldots,\;
k(\boldsymbol{\theta},\boldsymbol{\mu}_K)
\right]^\top \in \mathbb{R}^{K}.
\]

\paragraph{Nyström whitening (optional).}
Let $\mathbf{K}_{mm}\in\mathbb{R}^{K\times K}$ be the kernel matrix between centers,
$(\mathbf{K}_{mm})_{rs}=k(\boldsymbol{\mu}_r,\boldsymbol{\mu}_s)$.
\begin{revblock}
We optionally whiten features by
\[
\tilde{\boldsymbol{\phi}}(\boldsymbol{\theta})
= \mathbf{K}_{mm}^{-1/2}\boldsymbol{\phi}(\boldsymbol{\theta}),
\]
which is consistent with the column-vector definition of $\boldsymbol{\phi}(\boldsymbol{\theta})$ and corresponds to a Nyström approximation of the implicit feature map \cite{WilliamsSeeger2001}.
\end{revblock}

Finally, we fit a classical ridge regressor on the QKF feature vectors.

\begin{revblock}

\subsubsection{Variational quantum regressor (VQR)}
\label{subsec:vqr}
To retain the trainable quantum-circuit branch in the study while avoiding an enlarged primary model-selection budget, we also evaluated a compact variational quantum regressor as a supplementary analysis. The VQR used the same train-only compact angle-space inputs as QKR. Each input was encoded using the primary RY--RZ data-reuploading feature map, followed by one trainable entangling layer and Pauli-$Z$ measurements on each qubit. A classical linear readout mapped the measurement vector to a continuous prediction. The target was transformed with $\log(1+y)$ and standardised within the training fold, and all predictions were inverted to physical units before evaluation. VQR parameters were optimised with Adam using fixed training settings and fixed random seeds; no held-out fold information was used for preprocessing, target scaling, circuit training, or threshold definition. Because VQR introduces stochastic optimisation and trainable circuit parameters beyond the kernel ridge models, it is reported as a supplementary trainable-circuit baseline rather than as part of the primary repeated-CV compact-kernel ranking. The VQR implementation is available with the analysis code described in the Data and code availability statement.
\end{revblock}

\begin{revblock}

\subsection{SPD-kernel-embedded quantum regression and sensitivity analysis}
\label{subsec:spd_kernel_embedded_quantum}
\label{subsec:spd_kernel_embedded_lean_sensitivity}

To test whether SPD geometry could provide a more informative input representation for quantum-circuit-based regression, we evaluated an SPD-kernel-embedded quantum pipeline. This analysis differs from the standalone SPD prototype-divergence branch and from direct compact QKR. Instead of using the raw compact biomarkers directly as quantum angles, or simply appending SPD divergence features to the original input, the SPD geometry was first converted into a low-dimensional explicit embedding and then used as the quantum-model input.

 The reported analysis used a repeated cross-validation protocol with inner model selection. In each outer split, the outer training fold was used to construct the SPD descriptors, fit the SPD-kernel embedding, scale the quantum-circuit inputs, and select all model hyperparameters by inner cross-validation only. The held-out outer fold was evaluated once using the configuration selected within the corresponding outer-training data. No held-out outer-fold RMSE, MAE, \(R^2\), or ROC-AUC values were used to select the SPD kernel, embedding dimension, circuit setting, or regularisation parameter.

For each outer cross-validation split, SPD descriptors were constructed using training-fold preprocessing only. Let \(S_i \in \mathcal{S}_{++}^{p}\) denote the SPD matrix associated with sample \(i\). The primary SPD similarity was the Stein-kernel similarity. Pairwise divergences between training SPD matrices were computed using the Stein divergence,
\begin{equation}
D_{\mathrm{Stein}}(A,B)
=
\log \det \left( \frac{A+B}{2} \right)
-
\frac{1}{2}\log \det(A)
-
\frac{1}{2}\log \det(B),
\label{eq:stein_divergence_spd_embedding}
\end{equation}
for \(A,B \in \mathcal{S}_{++}^{p}\). The corresponding Stein-kernel similarity was
\begin{equation}
K_{\mathrm{Stein}}(S_i,S_j)
=
\exp \left[-\gamma D_{\mathrm{Stein}}(S_i,S_j)\right],
\label{eq:stein_kernel_spd_embedding}
\end{equation}
where \(\gamma > 0\) controls the similarity scale. The Stein-kernel similarity was computed using the training fold only. Positive-semidefiniteness was not assumed a priori for every finite training matrix and parameter setting; instead, each training-fold similarity matrix was symmetrised and checked numerically before explicit embedding.

As a targeted sensitivity analysis, we also evaluated a log-cosine angular SPD similarity. Each SPD descriptor was first mapped to the tangent vector space by the matrix logarithm and vectorised,
\begin{equation}
v(S) = \operatorname{vec}\{\log(S)\}.
\end{equation}
The log-cosine similarity between two SPD descriptors was then
\begin{equation}
K_{\mathrm{logcos}}(A,B)
=
\frac{\langle v(A),v(B)\rangle}
{\|v(A)\|_{2}\,\|v(B)\|_{2}+\varepsilon},
\label{eq:logcos_spd_kernel}
\end{equation}
where \(\varepsilon>0\) prevents numerical division by zero. The log-cosine analysis was included because a directional comparison in log-SPD coordinates can be competitive for compact SPD descriptors. It was not used as an exhaustive SPD-kernel search; it was a controlled comparison between the primary Stein geometry and one pre-specified angular SPD alternative. The log-cosine construction was treated as an angular SPD similarity rather than as a guaranteed positive-semidefinite kernel. If a finite similarity matrix was indefinite, only retained non-negative eigencomponents were used for explicit coordinates, and the corresponding analysis was interpreted as a regularised finite-similarity embedding rather than a strict Mercer-kernel model.

For either SPD similarity, the finite training-fold similarity matrix was symmetrised and, where necessary, numerically regularised. It was then eigendecomposed to obtain a low-dimensional explicit kernel embedding. Let
\begin{equation}
K_{\mathrm{SPD}}^{\mathrm{train}}
=
U \Lambda U^{\top}
\end{equation}
denote the eigendecomposition of the regularised training SPD similarity matrix, with eigenvalues sorted in descending order. For a chosen embedding dimension \(r \in \{1,2,3,4\}\), each training sample was represented by
\begin{equation}
z_i
=
\left[
\sqrt{\lambda_1}U_{i1},
\sqrt{\lambda_2}U_{i2},
\ldots,
\sqrt{\lambda_r}U_{ir}
\right]^{\top}
\in \mathbb{R}^{r}.
\label{eq:spd_kernel_embedding_coordinates}
\end{equation}
For a held-out test sample \(S_*\), out-of-sample coordinates were obtained by a Nystr\"om-style extension,
\begin{equation}
z_*
=
\left[
\frac{k_*^{\top}u_1}{\sqrt{\lambda_1}},
\frac{k_*^{\top}u_2}{\sqrt{\lambda_2}},
\ldots,
\frac{k_*^{\top}u_r}{\sqrt{\lambda_r}}
\right]^{\top},
\label{eq:nystrom_spd_embedding}
\end{equation}
where \(k_* = [K_{\mathrm{SPD}}(S_*,S_1),\ldots,K_{\mathrm{SPD}}(S_*,S_{n_{\mathrm{tr}}})]^{\top}\) contains similarities only to training-fold SPD matrices. Eigenvalues below numerical tolerance were excluded from the retained embedding; if an indefinite finite similarity matrix occurred, only the retained non-negative eigenspectrum was used for the explicit coordinates. The binary condition indicator \(c_i\) was appended to the SPD-kernel embedding,
\begin{equation}
\tilde{z}_i = [z_i^{\top}, c_i]^{\top}.
\label{eq:spd_embedding_condition_append}
\end{equation}

Two quantum-circuit-based regressors were evaluated on this representation. In the SPD-kernel-embedded QKR model, \(\tilde{z}_i\) was angle-scaled within the training fold and encoded into the same fidelity quantum-kernel ridge-regression framework used for the direct compact QKR baseline. The resulting kernel was
\begin{equation}
K_{\mathrm{QKR}}(\tilde{z}_i,\tilde{z}_j)
=
\left|
\left\langle
\psi(\theta(\tilde{z}_i))
\middle|
\psi(\theta(\tilde{z}_j))
\right\rangle
\right|^2,
\label{eq:spd_embedded_qkr_kernel}
\end{equation}
where \(\theta(\cdot)\) denotes train-only angle scaling and \(|\psi(\cdot)\rangle\) is the quantum state produced by the feature-map circuit.

In the SPD-kernel-embedded VQC model, the same embedded input \(\tilde{z}_i\) was passed through a shallow variational circuit. Pauli-\(Z\) expectation values were extracted from the circuit output and used as features for a regularised linear readout. Because this branch uses trainable circuit parameters and a classical readout rather than a fidelity kernel ridge system, it is reported separately from QKR and is referred to as SPD-kernel-embedded VQC rather than QKR.

To reduce model-selection pressure in the small cohort, the targeted analysis used a pre-specified low-dimensional hyperparameter grid. The SPD embedding dimension was selected from \(r\in\{1,2,3,4\}\). The SPD-kernel scale multiplier was fixed to 1.0, the quantum feature-map depth was fixed to one layer, and the angle-scaling values were selected from \(\{0.25,0.50\}\). For the SPD-kernel-embedded VQC branch, ridge readout regularisation was selected from \(\alpha\in\{0.1,1,10\}\), the circuit architecture was fixed to the RY--RZ CNOT-chain map, and the number of random VQC initialisation trials was fixed to four. For the SPD-kernel-embedded QKR branch, kernel-power rescaling was fixed to 1.0, and ridge regularisation was selected from \(\alpha\in\{0.1,1,10\}\). All selections were made using three-fold inner cross-validation inside each outer-training split. The outer evaluation used repeated stratified five-fold cross-validation with three repeats, giving 15 matched held-out folds per endpoint.

\begin{revblock}
\subsection{SPD-normalized ROSE/SteinDiv quantum bridge}
\label{subsec:spd_normalized_rose_bridge}

To test whether the SPD component provides a useful representation rather than only a descriptive label, we added an SPD-normalized geometry-to-quantum bridge inspired by random projection on SPD manifolds. The design follows the ROSE principle: SPD matrices are first compared using the Stein-kernel similarity, random hyperplanes are represented through training exemplars in the RKHS, and the resulting projection coefficients form ordinary Euclidean coordinates for downstream learning \cite{ref83}. In the proposed bridge, this kernel-induced SPD/RKHS mapping is performed before normalisation, so that the normalisation step acts on a geometry-conditioned representation rather than on raw biomarker coordinates. This ordering is motivated by prior work showing that quantum models are sensitive to how classical data are encoded: the data-encoding strategy affects the accessible function class of variational quantum models, while quantum-kernel bandwidth and input scaling influence the induced kernel spectrum, inductive bias, and generalisation behaviour \cite{schuld2021encoding,shaydulin2022bandwidth,canatar2023bandwidth}. These studies do not describe the present RKHS-normalised SPD bridge directly, but they support the broader principle that the scale and geometry of the representation supplied to a quantum feature map can materially affect model behaviour. The key difference from direct compact QKR is therefore that the quantum layer receives low-dimensional, train-normalised coordinates from the geometry-induced feature space rather than raw biomarker coordinates.

For each outer split, all operations were restricted to the corresponding training data. Continuous covariates were transformed using train-only preprocessing, compact SPD descriptors were constructed from the continuous biomarker representation, and synthetic SPD descriptors were generated from the outer-training fold only by log-Euclidean interpolation/extrapolation followed by train-derived radius control. The encoded condition indicator was not used in the SPD synthetic/reference construction for this bridge and was appended only after the geometry-induced representation was formed. Synthetic descriptors were unlabeled and were used only as geometric support/reference points; they were never used as target-labelled observations. The number of synthetic points was treated as a sensitivity parameter, with synthetic multipliers 0.5 and 1.0 corresponding to approximately 85--86 and 170--171 synthetic SPD descriptors per outer-training fold.

We evaluated three reference policies. First, the \emph{mixed-reference} policy used a random train-only reference set containing both real training SPD descriptors and training-derived synthetic SPD descriptors. Second, the \emph{synthetic-only} policy used only synthetic SPD descriptors as the geometric reference set. Third, the \emph{real-to-synthetic} audit represented each real sample only by Stein divergences or similarities to synthetic SPD descriptors, so no real-real divergence/similarity matrix entered the reported feature vector. These policies were included because the ROSE construction depends on the completeness and distribution of the reference points used to generate random hyperplanes, and synthetic points can help or hurt depending on whether they stabilize or distort the training distribution.

Two geometry-to-quantum variants were evaluated. In \textbf{ROSE-Norm-VQR}, the Stein-kernel similarity
\begin{equation}
K_{\mathrm{Stein}}(A,B)=\exp[-\gamma D_{\mathrm{Stein}}(A,B)]
\end{equation}
was computed against the train-only reference set. The same finite-similarity convention was used for Stein-kernel reference matrices in this bridge: matrices were symmetrised within each outer-training split, numerical spectra were checked when explicit coordinates were required, and only training-fold information was used to fit normalisation and projection operators. ROSE-style random hyperplanes were then formed in the Stein-kernel RKHS using a reference size of 96 and $t_{\mathrm{RP}}=24$ exemplars per hyperplane. The projected coordinates were standardized using the outer-training fold only, the binary condition indicator was appended, and the resulting $r+1$ inputs were encoded into a low-qubit VQR-style readout. In \textbf{SteinDiv-Hilbert-QKR}, the Stein divergence matrix was converted directly into a low-dimensional Hilbert/MDS embedding with Nystr\"om-style extension for held-out samples, followed by train-only normalization, condition appending, and QKR. For the full-mesh bridge analysis, the embedding dimension $r\in\{1,2,3,4\}$, angle scale $s\in\{0.25,0.50\}$, synthetic multiplier $\in\{0.5,1.0\}$, and ridge regularisation $\alpha\in\{0.1,1,10\}$ were selected by three-fold inner cross-validation inside each outer-training set when fold-level outputs were available. The compact VQR baseline used the same compact biomarker-plus-condition angle representation without the SPD random-projection stage. The outer evaluation used the same condition-stratified $3\times5$ repeated cross-validation protocol as the other main analyses.

Because the reference policy itself was a central sensitivity question, the main Results report the better ROSE reference policy for each endpoint and the Supplementary Information reports the complete mixed-reference, synthetic-only, and real-to-synthetic sensitivity tables. Corrected repeated-CV paired tests were computed for branches with fold-level outputs; aggregate-only sensitivity rows are reported descriptively and are not used for significance claims.

\end{revblock}

\subsection{Quantum-model ablation and sensitivity analysis}
\label{subsec:qkr_ablation_sensitivity}
To assess whether the quantum-model results were driven by specific design choices, we perform a dedicated ablation and sensitivity analysis. First, we evaluated pre-specified QKR feature-map configurations varying the number of data-reuploading layers ($L=1,2,3$), the angle-scale factor ($s=0.25,0.50,1.00$), and kernel-power rescaling ($\rho=0.5$ or $\rho=1.0$ where applicable). Ridge regularisation was selected by inner cross-validation within each outer-training split. Second, we varied the number of clustered quantum-feature centres in QKF ($K=3,6,9,12$), again using train-only angle-space preprocessing and inner-CV regularisation selection. Third, we evaluated feature-set sensitivity by comparing the manuscript-selected biomarkers plus condition, the same biomarkers without condition, leave-one-biomarker-out variants, all measured non-outcome predictors plus condition, and a condition-only control. Fourth, we perform a pre-specified encoding and circuit-design ablation. This compared the primary RY--RZ ring-CNOT angle map with RY-only angle encoding, RX--RZ angle encoding, phase RZ+CZ encoding, an IQP-style Z/ZZ phase map, and an amplitude-style normalised-state comparator. We also varied the entanglement topology of the primary RY--RZ map using no entanglement, linear CNOT, ring CNOT, and a full CNOT cascade. To isolate circuit architecture from additional hyperparameter search in the small cohort, all encoding/topology variants used pre-specified reference settings ($L=2$, $s=0.5$, kernel power $\rho=1.0$, and ridge regularisation $\alpha=10$). Because different encodings can produce kernels with different diagonal and off-diagonal scales, this common-$\alpha$ design is a scale-sensitive architecture screen rather than a fully tuned ranking of encoding families; it may disadvantage encodings whose optimal regularisation differs from the primary angle map. To keep this screen conservative, we did not use the pre-specified-encoding table to select an alternative circuit. A supplementary regularisation-sensitivity check tunes ridge regularisation separately for each encoding. This check is reported as a sensitivity audit only, because it adds another model-selection dimension beyond the pre-specified-reference comparison.

All quantum-ablation analyses used the same repeated stratified five-fold outer splits as the repeated-CV robustness analysis, with stratification by experimental condition and all preprocessing, feature scaling, PCA/angle mapping, centre selection, and hyperparameter selection performed within the training data only. The purpose of this analysis was not to exhaustively optimise the quantum model, but to test whether the reported behaviour was stable to nearby feature-map choices, clustered-centre choices, feature-selection choices, and fold variation.
\end{revblock}

\begin{revblock}
\subsection{Small-sample hyperparameter control for quantum models}
\label{subsec:small_sample_quantum_control}
Quantum-kernel models introduce several design parameters, including feature-map depth, angle scaling, kernel-power rescaling, ridge regularisation, and the number of QKF centres. Because the cohort contains $n=213$ animals, each stratified five-fold outer split has approximately $n_{\mathrm{train}}\approx170$ and $n_{\mathrm{test}}\approx43$ samples. We therefore separated hyperparameter control from held-out evaluation. The outer held-out fold was used only for held-out metric estimation. Within each outer training split, preprocessing, PCA/angle mapping where required, centre selection, ridge-penalty selection, kernel hyperparameter selection, and screening-threshold definition were fit without access to the held-out fold. The quantum feature-map sensitivity grid was then reported as a stability analysis rather than as a post hoc search for an alternative best model.

To reduce validation-set overfitting, we used compact, pre-specified grids and interpreted sensitivity ranges rather than selecting a single grid point as definitive. For the tuned QKR model used in the main compact comparison, the prespecified grid contained two data-reuploading depths ($L=1,2$), four angle-scale values ($s=0.10,0.25,0.50,1.00$), two kernel-power settings ($\rho=1.0,0.5$), and eight ridge-regularisation values. This corresponds to 16 quantum feature-map configurations and 128 total QKR candidate settings when ridge penalties are counted. For QKF, the grid contained three centre counts ($K=3,6,9$) and seven ridge penalties, giving 21 candidate settings. \textit{These candidate counts describe the train-only tuning budget, not additional test-fold evaluations.} We therefore interpret the selected hyperparameters as stable candidates from pre-specified grids, not as biologically or algorithmically definitive optima. The pre-specified feature-map and centre-count sensitivity analyses quantify whether conclusions are stable to nearby quantum-design choices, but they do not remove the fundamental limitation that hyperparameter assessment in a cohort of this size remains uncertain and requires independent validation.
\end{revblock}

 \subsection{Evaluation metrics}
\label{subsec:metrics}

\subsubsection{Regression metrics}
Let $y_i$ denote the true target in physical units and $\hat{y}_i$ the prediction
(in physical units after inverting any target transform).
We report:
\[
\mathrm{RMSE} = \sqrt{\frac{1}{m}\sum_{i=1}^{m}(y_i-\hat{y}_i)^2}, \quad
\]
\begin{revblock}
and mean absolute error,
\[
\mathrm{MAE} = \frac{1}{m}\sum_{i=1}^{m}|y_i-\hat{y}_i|,
\]
\end{revblock}
and coefficient of determination
\[
R^2 = 1 - \frac{\sum_{i=1}^{m}(y_i-\hat{y}_i)^2}{\sum_{i=1}^{m}(y_i-\bar{y})^2}.
\]
We also report a scale-normalised percent error for comparison across endpoints:
\[
\%{\rm RMSE} = 100\cdot \frac{{\rm RMSE}}{\overline{y}_{\rm test}}, \qquad
{\rm } \]
where $\overline{y}_{\rm test}$ is the mean target value in the held-out portion of a given outer fold. Reported percent errors are averaged across the relevant outer-fold evaluations.

\subsubsection{Screening metrics derived from regression outputs}
To interpret regression models as screening tools, we define a binary label
$\ell_i\in\{0,1\}$ from the continuous target using a threshold $\tau$ computed from Sham subjects in the training split. For ``low'' screening:
\[
\ell_i =
\begin{cases}
1, & y_i \le \tau,\\
0, & y_i > \tau.
\end{cases}
\]
 In the reported experiments, $\tau$ is defined as 0.8 times the Sham-group mean computed on the training portion of each outer fold.

 Given regression predictions $\hat{y}_i$, we evaluate screening performance using the continuous score $s_i=-\hat{y}_i$, so that larger scores indicate stronger evidence for  the positive class. ROC-AUC is then computed on the held-out outer fold.

\begin{revblock}
\subsubsection{Descriptive screening-calibration audit}
\label{subsec:screening_calibration_audit}
We perform a secondary descriptive calibration audit for the derived low-outcome screening task. The event definition used the same biologically defined threshold as the ROC-AUC analysis: within each outer-training fold, an event was defined as an observed endpoint value below 80\% of the training-fold Sham mean. Because the primary models are continuous regressors rather than native probability models, continuous predictions were converted to low-outcome risk probabilities using a train-only logistic recalibration model fitted to the score $\tau-\hat{y}_i$, where $\tau$ is the training-fold threshold. The fitted calibrator was then applied once to the held-out outer test fold. We report Brier score, five-bin expected calibration error (ECE), calibration slope, and event prevalence as descriptive screening-calibration metrics. This audit is interpreted descriptively because the endpoint is a derived screening label in a small preclinical cohort, not a validated clinical diagnosis or decision threshold.
\end{revblock}

\begin{revblock}
\subsection{Statistical validation and uncertainty quantification}
\label{subsec:stat_validation}
To quantify uncertainty in the small-sample performance estimates, we perform a repeated cross-validation robustness analysis for the compact classical and quantum comparison. We used repeated stratified five-fold cross-validation with three repeats, stratifying by experimental condition, so that each model was evaluated on identical held-out folds. For each outer split, median imputation, Yeo--Johnson transformation, standardisation, PCA-to-qubit reduction where applicable, angle scaling, SPD prototype construction where applicable, and model-parameter selection were repeated from the corresponding training partition before evaluation on the matched held-out fold.

For each model and endpoint, the main tables report mean held-out-fold performance and fold-level RMSE standard deviation. Full unreduced tables and complementary metric summaries are provided in the Supplementary Information. Pairwise model comparisons were performed using matched fold-level RMSE differences between the best quantum/SPD model and the best classical comparator for each endpoint. The primary paired comparison used the Wilcoxon signed-rank test, with Holm adjustment across the three endpoint-level comparisons \cite{holm1979simple}. Because repeated cross-validation folds share observations and are not statistically independent, these $p$-values were interpreted conservatively and were used to assess whether numerical differences were robust rather than to establish external-validity claims. Corrected repeated-CV paired $t$-tests using the Nadeau--Bengio correction factor $1/(Rk)+n_{\mathrm{test}}/n_{\mathrm{train}}$, where $R$ is the number of repeats and $k=5$ is the number of folds, were reported as a parametric sensitivity analysis in the Supplementary Information \cite{nadeau2003inference}.
\end{revblock}

\subsection{Implementation}
All models were implemented in Python.
Classical models used standard implementations from scikit-learn.
\begin{revblock}
Quantum kernels were evaluated using classical statevector simulation with the PennyLane simulator to compute fidelities exactly. The supplementary VQR used the same statevector-circuit specification and was optimised with deterministic differentiable tensor operations. Statevector simulation provides a controlled setting for evaluating quantum feature-map behaviour independently of sampling noise and device calibration. Physical-device execution, noise-aware hardware evaluation, resource-scaling analysis, and runtime-speed comparison are outside the scope of this study.
\end{revblock}


\begin{revblock}

\end{revblock}

\subsection{Compared model families}
\begin{revblock}
Classical references included global and condition-mean baselines where available, Ridge regression on raw and engineered features, random forests, shallow decision trees, LDA condition-axis Ridge, compact Ridge, angle-space Ridge, RBF and polynomial kernel-ridge regression, and classical SPD-divergence-feature or SPD-similarity models. Similar variants from the same family are reduced in the main tables by retaining the lower-RMSE representative, while the unreduced tables are reported in the Supplementary Information.

SPD-divergence-feature models converted compact biomarker vectors into SPD descriptors and compared them using Stein divergence or log-domain angular similarity. Quantum models used shallow angle-encoding circuits and either full fidelity-kernel ridge regression, clustered/Nystr\"om quantum features, or VQC-style readouts. SPD-kernel-embedded models first converted the SPD similarity matrix into a low-dimensional coordinate representation, appended condition, and then used QKR or VQC-style readouts. ROSE/Synth\_ROSE models generated training-fold synthetic SPD matrices, constructed a Stein-kernel random-projection or Hilbert-space representation, normalised after the SPD/RKHS step, and supplied the resulting coordinates to a low-qubit quantum readout. The real-to-synthetic audit represented real samples only through Stein divergences or similarities to synthetic SPD references generated inside the outer-training fold.
\end{revblock}

\begin{revblock}
\subsection{Evaluation and statistical testing}
RMSE was the primary ranking metric. MAE, $R^2$, and ROC-AUC were reported as complementary metrics. ROC-AUC was computed from continuous predictions after defining a training-fold Sham-referenced low-outcome threshold. Main-text rankings use the repeated-CV results from the 15 held-out outer-fold evaluations. For paired inference, matched fold-level RMSE differences were compared between the best quantum/SPD model and the best classical model for each endpoint. The primary paired test was the Wilcoxon signed-rank test with Holm adjustment across the three endpoint-level comparisons; a corrected repeated-CV paired $t$-test is reported as a parametric sensitivity statistic in the Supplementary Information.
\end{revblock}

\section{Results}
\label{sec:results}

\subsection{Tibialis anterior muscle weight}
\begin{revblock}
For muscle weight, Synth\_ROSE had the numerically lowest mean RMSE (5.116 mg; Table~\ref{tab:weight_main_final}), improving numerically on the best classical row, RBF-KRR (5.210 mg), and on the best SPD-kernel-embedded VQC row (5.159 mg). The best MAE in the displayed table was obtained by QKR-full direct compact (3.401 mg), illustrating that endpoint ranking can depend on the error metric even when RMSE is used as the primary criterion.
\end{revblock}

\begin{table}[!ht]
\centering
\scriptsize
\caption{Repeated 3$\times$5 cross-validation results for tibialis anterior muscle weight. Classical models are listed first, followed by quantum/SPD geometry-aware models.}
\label{tab:weight_main_final}
\resizebox{\textwidth}{!}{%
\begin{tabular}{llrrrrrr}
\toprule
Group & Model & CV folds & RMSE $\downarrow$ & RMSE SD & MAE $\downarrow$ & $R^2$ $\uparrow$ & ROC--AUC $\uparrow$ \\
\midrule
Classical & RBF-KRR & 15 & 5.210 & 0.770 & 3.432 & 0.473 & 0.786 \\
Classical & Angle-space Ridge & 15 & 5.231 & 0.671 & 3.487 & 0.469 & 0.752 \\
Classical & Compact Ridge + condition & 15 & 5.262 & 0.667 & 3.520 & 0.462 & 0.750 \\
Classical & Original compact Ridge & 15 & 5.289 & 0.658 & 3.553 & 0.456 & 0.756 \\
Classical & Ridge + SPD divergences, Stein & 15 & 5.290 & 0.739 & 3.557 & 0.457 & 0.766 \\
Classical & Biomarker-only Ridge & 15 & 5.368 & 0.624 & 3.680 & 0.439 & 0.763 \\
Classical & Quadratic Ridge + condition & 15 & 5.397 & 0.761 & 3.542 & 0.432 & 0.777 \\
Classical & SPD-KRR only, Stein & 15 & 6.710 & 0.741 & 5.147 & 0.127 & 0.677 \\
\midrule
Quantum/SPD & Synth\_ROSE & 15 & \textbf{5.116} & 0.705 & 3.422 & \textbf{0.492} & 0.768 \\
Quantum/SPD & Stein SPD embedded VQC, r=1 & 15 & 5.159 & 0.645 & 3.407 & 0.483 & 0.780 \\
Quantum/SPD & ROSE & 15 & 5.174 & 0.710 & 3.448 & 0.480 & 0.775 \\
Quantum/SPD & QKR-full direct compact & 15 & 5.184 & 0.714 & \textbf{3.401} & 0.475 & 0.788 \\
Quantum/SPD & Logcosine SPD embedded VQC, r=1 & 15 & 5.217 & 0.725 & 3.448 & 0.471 & 0.773 \\
Quantum/SPD & SteinDiv-Hilbert-QKR & 15 & 5.506 & 0.798 & 3.875 & 0.415 & 0.781 \\
\bottomrule
\end{tabular}%
}
\vspace{0.25em}
\begin{minipage}{0.98\textwidth}\footnotesize Values are means over 15 held-out folds from repeated stratified five-fold cross-validation. The column ``CV folds'' refers to the 15 outer held-out evaluations from 3$\times$5 repeated cross-validation, not to animal sample size. Rows retain the lower-RMSE representative when two variants from the same family had very similar performance. The complete unreduced tables are provided in the Supplementary Information.\end{minipage}
\end{table}

\subsection{Tibialis anterior force readout}
\begin{revblock}
For force readout, the lowest RMSE was obtained by the best classical row, Biomarker-only Ridge (2358.459 mN; Table~\ref{tab:force_main_final}). The best quantum/SPD row by RMSE was QKR-full direct compact (2369.948 mN), closely followed by QKF-cluster (2370.610 mN). Thus, the force endpoint favours a strong compact classical model, although quantum/SPD rows remain close in absolute RMSE.
\end{revblock}

\begin{table}[!ht]
\centering
\scriptsize
\caption{Repeated 3$\times$5 cross-validation results for tibialis anterior force readout. Classical models are listed first, followed by quantum/SPD geometry-aware models.}
\label{tab:force_main_final}
\resizebox{\textwidth}{!}{%
\begin{tabular}{llrrrrrr}
\toprule
Group & Model & CV folds & RMSE $\downarrow$ & RMSE SD & MAE $\downarrow$ & $R^2$ $\uparrow$ & ROC--AUC $\uparrow$ \\
\midrule
Classical & Global mean baseline & 15 & 3619.197 & 506.810 & 2977.348 & -0.016 & 0.500 \\
Classical & Condition means baseline & 15 & 2388.222 & 214.474 & 2018.753 & 0.547 & 0.856 \\
Classical & Biomarker-only Ridge & 15 & \textbf{2358.459} & 233.420 & \textbf{1981.346} & \textbf{0.562} & \textbf{0.896} \\
Classical & LDA condition-axis Ridge, engineered & 15 & 2364.686 & 256.495 & 1987.384 & 0.560 & 0.888 \\
Classical & Compact Ridge + condition & 15 & 2365.346 & 232.737 & 1989.088 & 0.559 & 0.894 \\
Classical & Ridge + SPD divergences, Stein & 15 & 2377.302 & 213.438 & 1999.807 & 0.553 & 0.884 \\
Classical & RBF-KRR & 15 & 2382.464 & 232.329 & 2003.678 & 0.553 & 0.892 \\
Classical & Random forest, raw & 15 & 2422.689 & 239.245 & 2056.354 & 0.536 & 0.873 \\
Classical & Shallow decision tree, engineered & 15 & 2485.297 & 238.731 & 2080.825 & 0.507 & 0.847 \\
Classical & SPD-KRR only, Stein & 15 & 3073.967 & 381.536 & 2430.282 & 0.258 & 0.805 \\
\midrule
Quantum/SPD & QKR-full direct compact & 15 & 2369.948 & \textbf{232.293} & 1992.915 & 0.557 & 0.886 \\
Quantum/SPD & QKF-cluster, Nystrom m=3 & 15 & 2370.610 & 239.402 & 1997.355 & 0.558 & 0.883 \\
Quantum/SPD & SteinDiv-Hilbert-QKR & 15 & 2381.997 & 205.386 & 2006.461 & 0.550 & 0.877 \\
Quantum/SPD & Stein SPD embedded QKR, r=1 & 15 & 2398.519 & 169.856 & 2030.747 & 0.540 & 0.883 \\
Quantum/SPD & ROSE & 15 & 2408.157 & 265.347 & 2020.802 & 0.543 & 0.847 \\
\bottomrule
\end{tabular}%
}
\vspace{0.25em}
\begin{minipage}{0.98\textwidth}\footnotesize Values are means over 15 held-out folds from repeated stratified five-fold cross-validation. The column ``CV folds'' refers to the 15 outer held-out evaluations from 3$\times$5 repeated cross-validation, not to animal sample size. Rows retain the lower-RMSE representative when two variants from the same family had very similar performance. The complete unreduced tables are provided in the Supplementary Information.\end{minipage}
\end{table}

\subsection{Muscle quality index}
\begin{revblock}
For muscle quality, Synth\_ROSE had the numerically lowest mean RMSE (58.249 mN/mg; Table~\ref{tab:quality_main_final}), slightly below the best classical row, Original compact Ridge (58.380 mN/mg), and the log-cosine SPD embedded VQC model (58.291 mN/mg). The absolute difference between Synth\_ROSE and the best classical row was 0.131 mN/mg, so the result is best interpreted as a small endpoint-specific numerical gain rather than a large separation.
\end{revblock}

\begin{table}[!ht]
\centering
\scriptsize
\caption{Repeated 3$\times$5 cross-validation results for the muscle quality index. Classical models are listed first, followed by quantum/SPD geometry-aware models.}
\label{tab:quality_main_final}
\resizebox{\textwidth}{!}{%
\begin{tabular}{llrrrrrr}
\toprule
Group & Model & CV folds & RMSE $\downarrow$ & RMSE SD & MAE $\downarrow$ & $R^2$ $\uparrow$ & ROC--AUC $\uparrow$ \\
\midrule
Classical & Original compact Ridge & 15 & 58.380 & 5.948 & 47.448 & 0.225 & 0.722 \\
Classical & Biomarker-only Ridge & 15 & 58.470 & 5.918 & \textbf{47.406} & 0.223 & 0.723 \\
Classical & Compact Ridge + condition & 15 & 58.640 & 5.929 & 47.570 & 0.218 & 0.722 \\
Classical & Ridge + SPD angular similarity & 15 & 58.695 & 5.635 & 47.577 & 0.216 & 0.715 \\
Classical & Angle-space Ridge & 15 & 58.741 & 5.969 & 47.714 & 0.215 & 0.719 \\
Classical & Polynomial-KRR & 15 & 58.829 & 5.951 & 47.717 & 0.213 & 0.723 \\
Classical & Quadratic Ridge + condition & 15 & 59.445 & 5.575 & 48.246 & 0.194 & 0.720 \\
Classical & SPD-KRR only, Stein & 15 & 62.714 & 6.165 & 49.951 & 0.106 & 0.678 \\
\midrule
Quantum/SPD & Synth\_ROSE & 15 & \textbf{58.249} & 5.615 & 47.867 & \textbf{0.227} & 0.715 \\
Quantum/SPD & Logcosine SPD embedded VQC, r=1 & 15 & 58.291 & 5.939 & 47.791 & 0.226 & \textbf{0.727} \\
Quantum/SPD & ROSE & 15 & 58.504 & 5.964 & 47.863 & 0.217 & 0.714 \\
Quantum/SPD & Stein SPD embedded VQC, r=1 & 15 & 58.716 & 7.021 & 48.167 & 0.215 & 0.723 \\
Quantum/SPD & SteinDiv-Hilbert-QKR & 15 & 58.734 & 6.051 & 48.010 & 0.215 & 0.713 \\
Quantum/SPD & QKR-full direct compact & 15 & 58.885 & 6.136 & 48.130 & 0.208 & 0.718 \\
\bottomrule
\end{tabular}%
}
\vspace{0.25em}
\begin{minipage}{0.98\textwidth}\footnotesize Values are means over 15 held-out folds from repeated stratified five-fold cross-validation. The column ``CV folds'' refers to the 15 outer held-out evaluations from 3$\times$5 repeated cross-validation, not to animal sample size. Rows retain the lower-RMSE representative when two variants from the same family had very similar performance. The complete unreduced tables are provided in the Supplementary Information.\end{minipage}
\end{table}

\subsection{Sensitivity analysis and paired statistical tests}
\begin{revblock}
Figure~\ref{fig:sensitivity_final} visualises fold-level ROC--AUC variability for four representative models across the three endpoints: QKR, Synth-ROSE, Biomarker-only Ridge, and RBF-KRR. The RMSE-based model ranking remains reported in Tables~\ref{tab:weight_main_final}--\ref{tab:quality_main_final}; ROC--AUC is shown as a complementary screening-oriented metric rather than as the primary ranking criterion. The figure shows that discrimination performance varies across folds and endpoints, reinforcing the need to interpret ROC--AUC together with the primary RMSE results and the paired repeated-CV statistical tests.

In the primary paired best-quantum/SPD versus best-classical comparison, force favoured the classical model (QKR-full minus Biomarker-only Ridge $\Delta$RMSE = +11.489 mN; Wilcoxon $p=0.890$, Holm-adjusted $p=1.000$). Muscle quality favoured Synth\_ROSE numerically (Synth\_ROSE minus Original compact Ridge $\Delta$RMSE = -0.131 mN/mg; Wilcoxon $p=0.762$, Holm-adjusted $p=1.000$). Muscle weight also favoured Synth\_ROSE numerically (Synth\_ROSE minus RBF-KRR $\Delta$RMSE = -0.094 mg; Wilcoxon $p=0.095$, Holm-adjusted $p=0.284$). Therefore, the statistical interpretation is conservative: the SPD-normalised synthetic-reference branch is numerically favourable for weight and quality, but statistical superiority is not established.
\end{revblock}

\section{Discussion}
\begin{revblock}
    
This study evaluated classical machine learning, SPD-geometry-based representations, and simulated quantum-circuit-based models for predicting three related but biologically distinct skeletal-muscle endpoints in a preclinical COPD dataset. All main-text model rankings were based on a single repeated cross-validation reference set, with train-only preprocessing, inner-fold model selection, and held-out folds used only for evaluation. The resulting interpretation is endpoint-specific rather than based on a single overall ranking. Strong classical models remained difficult to outperform, particularly for force readout, but the proposed kernel-geometric quantum bridge had the numerically lowest mean RMSE for muscle weight and muscle quality. This suggests that SPD/RKHS normalisation and compact quantum-kernel representations can provide useful predictive structure when the endpoint depends on richer biomarker relationships.

\begin{figure}[p]
  \centering

  \begin{minipage}{0.78\linewidth}
    \centering
    \includegraphics[width=0.7\linewidth]{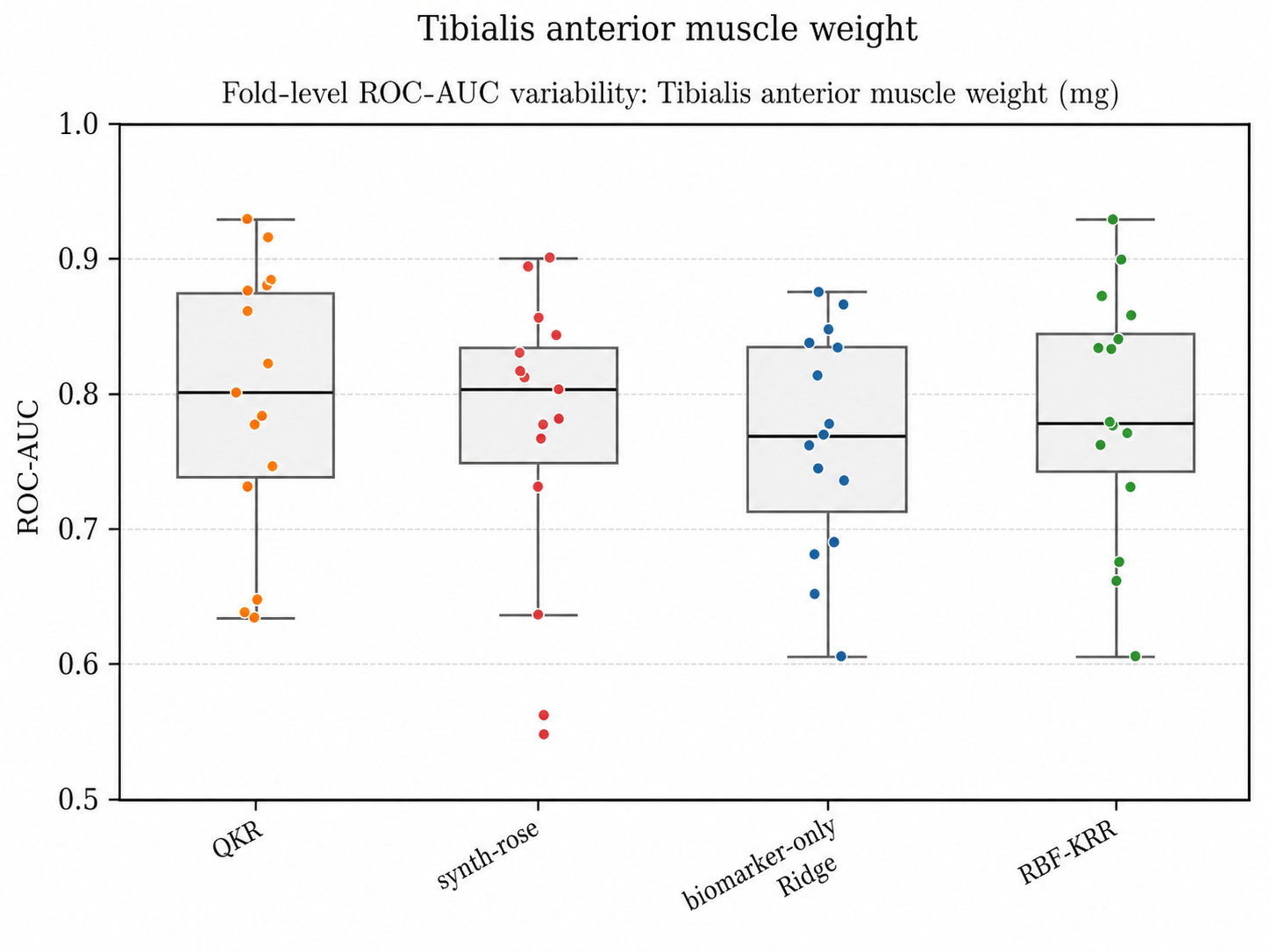}
  \end{minipage}

  \vspace{1.0em}

  \begin{minipage}{0.78\linewidth}
    \centering
    \includegraphics[width=0.7\linewidth]{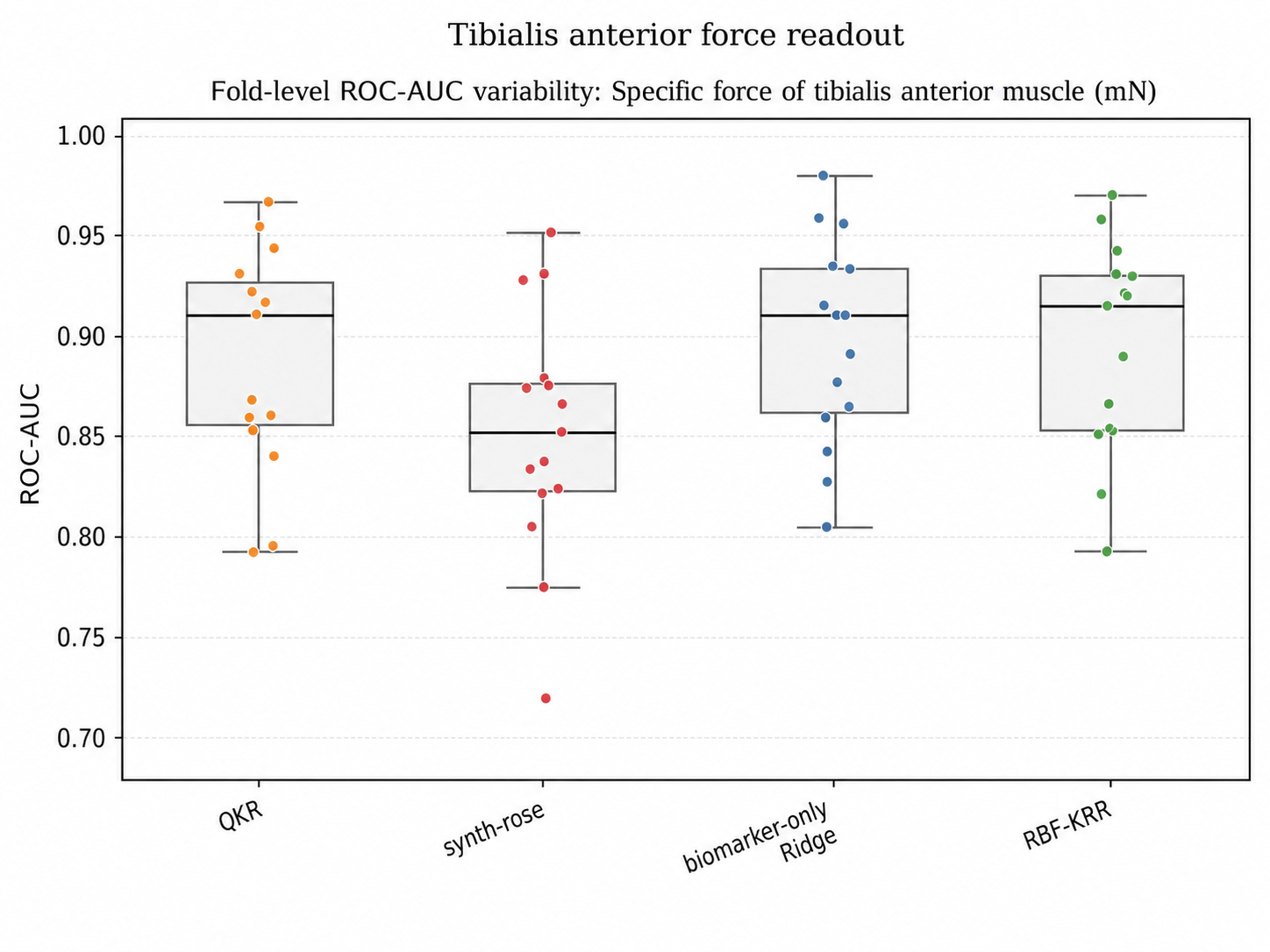}
  \end{minipage}

  \vspace{1.0em}

  \begin{minipage}{0.78\linewidth}
    \centering
    \includegraphics[width=0.7\linewidth]{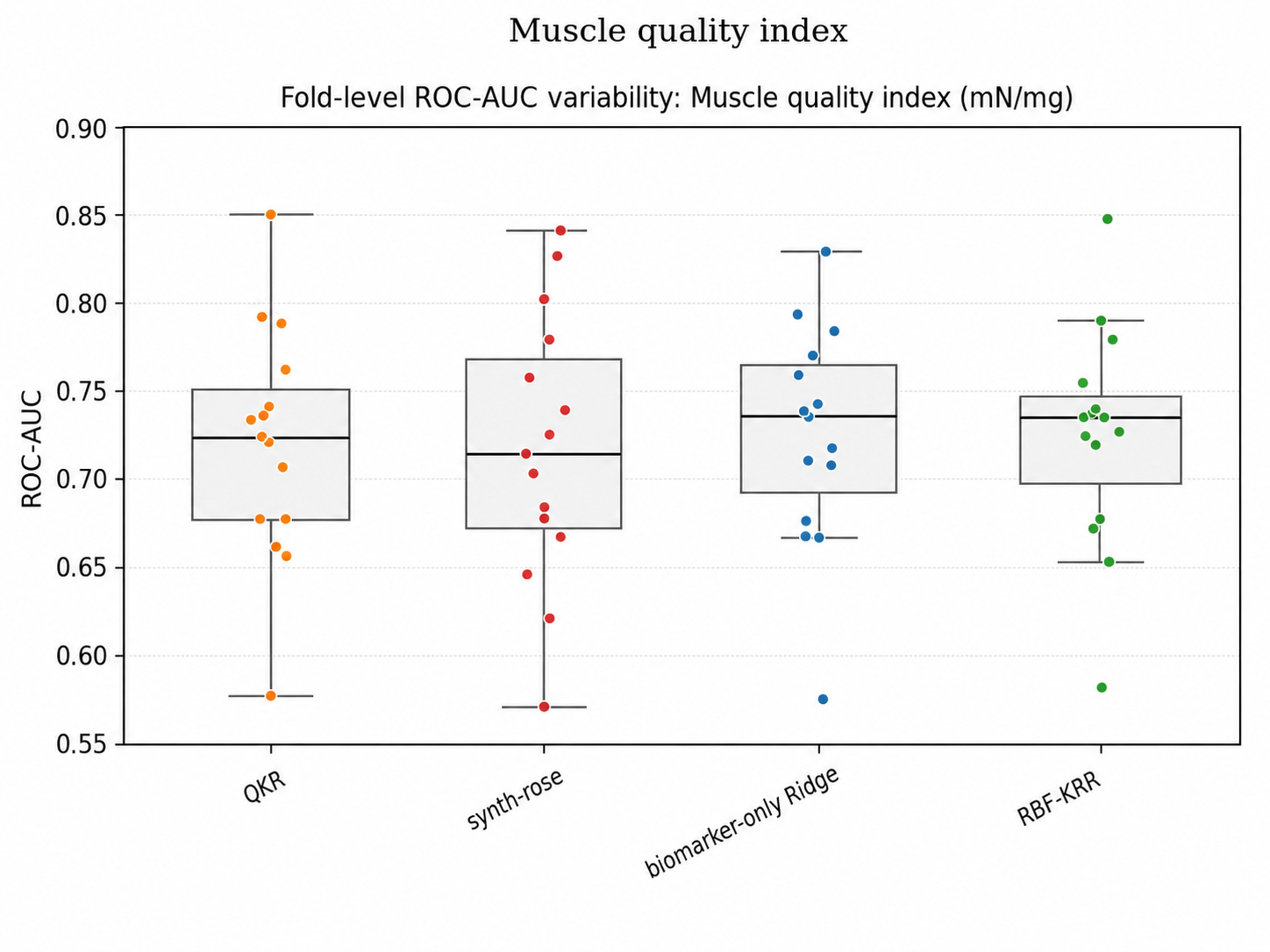}
  \end{minipage}

  \caption{Fold-level ROC-AUC variability for the repeated cross-validation analysis across the three skeletal-muscle endpoints. Panel from top to bottom: the top panel shows tibialis anterior muscle weight; the middle panel shows tibialis anterior force readout; and the bottom panel shows muscle quality index. Within each panel, methods are displayed from left to right in the same order: QKR, Synth-ROSE, Biomarker-only Ridge, and RBF-KRR. Point colours follow the same order: orange for QKR, red for Synth-ROSE, blue for Biomarker-only Ridge, and green for RBF-KRR. Each point corresponds to the ROC-AUC obtained on one held-out outer fold, giving 15 fold-level values per method. Boxes show the interquartile range, horizontal lines show the median, and whiskers show fold-level dispersion. These plots visualise fold-to-fold discrimination variability and complement the RMSE-centred repeated-cross-validation summary tables.}
  \label{fig:sensitivity_final}
\end{figure}

The repeated-CV results show clear endpoint dependence. For tibialis anterior force, Biomarker-only Ridge had the lowest mean RMSE (2358.459 mN), while the best quantum/SPD model, QKR-full direct compact, remained closely competitive but did not improve the primary RMSE criterion. This suggests that force readout in the present cohort may be dominated by a comparatively linear biomarker signal. In contrast, for muscle quality, Synth\_ROSE had the numerically lowest mean RMSE (58.249 mN/mg), narrowly improving on Original compact Ridge (58.380 mN/mg) and log-cosine SPD-embedded VQC (58.291 mN/mg). The clearest numerical gain was observed for muscle weight, where Synth\_ROSE had the numerically lowest mean RMSE (5.116 mg), improving on RBF-KRR (5.210 mg), Stein SPD-embedded VQC (5.159 mg), mixed-reference ROSE (5.174 mg), and QKR-full direct compact (5.184 mg). These findings show that the proposed representation is not simply adding model complexity; rather, it appears most useful when the target endpoint benefits from a geometry-aware representation of biomarker covariance structure. In less complex or more linearly structured settings, such as force in this cohort, the quantum/SPD models performed competitively but did not displace the strongest classical baseline. This pattern is also consistent with the mathematical definitions of the endpoints: force is a direct functional readout, muscle weight reflects tissue mass and may have a more complex nonlinear relationship with inflammatory and physiological biomarkers, and muscle quality is a derived ratio, $y^{(\mathrm{q})}=y^{(\mathrm{f})}/y^{(\mathrm{w})}$, combining information from both force and weight. Under this interpretation, a smaller gain for muscle quality is plausible because the endpoint inherits both the comparatively linear component seen for force and the richer covariance/nonlinear component suggested for weight.

The SPD component has a concrete representational role. It converts compact biomarker profiles into positive-definite descriptors that encode second-order relationships, then compares those descriptors using Stein divergence, log-domain angular similarity, ROSE-style RKHS random projection, or Hilbert/MDS coordinates. This is not merely an additional label for preprocessing; it changes the geometry in which biomarker interactions are measured before the low-qubit quantum readout. The central methodological question addressed in the analysis is whether kernel-induced SPD representations can be made more useful by normalising them in a reproducing-kernel Hilbert space before low-dimensional projection and quantum regression. Synth\_ROSE directly tests this idea. Synthetic SPD references are generated inside each outer-training fold, real samples are represented through Stein-kernel or Stein-divergence relationships to these references, the induced representation is normalised in RKHS feature space, and compact coordinates are supplied to a low-dimensional quantum regression model. This design separates the geometric reference construction from the held-out test fold and avoids using test information during synthetic generation, embedding, or normalisation.

The results are consistent with RKHS normalisation improving prediction when the endpoint benefits from geometric or covariance-based structure. Its largest numerical effect appeared for muscle weight, where Synth\_ROSE had the numerically lowest mean RMSE among all main-table models, and a smaller numerical gain was also observed for muscle quality. In contrast, force remained best predicted by a linear biomarker model. This endpoint-dependent behaviour is important because it avoids an overgeneral claim of quantum or SPD superiority while still demonstrating the value of the proposed representation. The fact that Synth\_ROSE was numerically lowest for two endpoints supports the usefulness of the SPD/RKHS bridge as a candidate representation, while the force endpoint shows that SPD geometry should still be judged against strong classical baselines.

The largest numerical difference was observed for muscle weight, where Synth\_ROSE had RMSE approximately 1.8\% lower than the best classical comparator. This finding should be interpreted as a repeated-CV preclinical signal, not as a claim of immediate clinical deployment or definitive model superiority. The paired tests did not establish statistical significance after Holm correction, and a relative reduction in regression error should not be converted directly into an absolute change in clinical risk-stratification accuracy. However, the signal remains biologically meaningful because the predicted phenotype is functionally relevant: skeletal-muscle dysfunction in COPD impairs exercise tolerance, daily function, independence, and quality of life, and strength loss may emerge before overt atrophy \cite{maltais2014limb,ref25,ref27}. Therefore, if similar improvements were reproduced in independent human cohorts with validated clinical thresholds, even modest predictive gains could have practical value for identifying individuals who may benefit from closer monitoring or earlier intervention.

This population-scale context is important because COPD affects hundreds of millions of people globally. A recent global modelling study estimated approximately 480 million COPD cases worldwide in adults aged at least 40 years \cite{boers2023global}. The present mouse regression result is not a diagnostic or trial-based estimate of human benefit, and it should not be used to calculate a specific number of people reclassified. Rather, the global burden explains why small, reproducible, externally validated improvements in biologically meaningful muscle-outcome prediction could still matter in future translational risk-stratification work.

A notable biological implication of the  results is that muscle weakness, muscle weight, and muscle quality appear contextually dissociable rather than interchangeable in this controlled preclinical COPD model. Strength deficits can arise without proportionate gross atrophy through oxidative stress, impaired calcium handling, mitochondrial dysfunction, vascular changes, and excitation--contraction uncoupling \cite{ref27,ref49,ref52,maltais2014limb}. The main tables reinforce this distinction: the best force model differs from the best weight and quality models. Thus, force may be a sensitive indicator of functional impairment, while muscle weight and muscle quality may reflect different aspects of systemic and tissue-level disease burden. The results support a shift from viewing COPD muscle dysfunction only as a mass-loss problem toward recognising weakness and muscle quality as distinct and functionally important manifestations.

This distinction has translational relevance because human COPD is biologically heterogeneous. Inflammation, disuse, exacerbation history, nutritional status, comorbidities, and disease endotype may contribute differently to muscle phenotypes. In the present cohort, force was best predicted by a linear biomarker-only ridge model, suggesting that this endpoint may be more directly explained by the available biomarker signal. Muscle weight and muscle quality, by contrast, benefited from the proposed kernel-geometric quantum representation, suggesting that these endpoints may contain richer covariance structure that is not fully captured by a simple linear model. This endpoint-specific pattern strengthens, rather than weakens, the contribution of the study: it shows where geometry-aware quantum representations appear useful, and where simpler classical models remain sufficient.

The quantum models in this study are simulator-based feature maps and kernel regressors. They do not establish quantum computational advantage, hardware speed-up, or clinical readiness. Their value here is methodological: they provide controlled nonlinear similarity maps that can be compared with RBF, polynomial, SPD, and ROSE-based alternatives under the same small-data protocol. Direct QKR, clustered/Nyström quantum-kernel features, SPD-kernel-embedded circuits, VQR variants, and the Synth\_ROSE bridge were evaluated against strong classical ridge and kernel baselines. This comparison is important because small biomedical datasets can make flexible models appear promising unless they are benchmarked against well-regularised classical methods. The strongest supported statement is therefore endpoint-dependent numerical competitiveness, with Synth\_ROSE providing the best RMSE for muscle weight and muscle quality and QKR-based models remaining close to the strongest classical alternatives for force.

All repeated-CV analysis results were obtained under repeated cross-validation with train-only preprocessing and model selection restricted to the training portion of each outer split. Synthetic SPD samples were generated only within the training fold and were unlabeled; they were used to define geometric reference systems rather than to add target information. The sensitivity analyses further examined the stability of the conclusions across SPD embedding dimension, ROSE versus Synth\_ROSE reference policy, real-to-synthetic reference construction, and compact quantum variants. These analyses did not overturn the main endpoint-specific interpretation. Instead, they strengthen the methodological conclusion that the SPD/RKHS bridge can be implemented in a leakage-controlled manner and that its benefit is most apparent for muscle weight and, to a lesser extent, muscle quality.

This remains a small preclinical study. The dataset contains 213 animals, and repeated-CV folds are not statistically independent despite repeated evaluation and paired fold-level testing. The paired Wilcoxon signed-rank tests and Holm adjustment provide a more cautious comparison than mean performance alone, but they do not replace independent validation. The quantum models were evaluated by statevector simulation, so the results concern feature-map behaviour, not quantum hardware speed-up, noise-resilient execution, or computational advantage. The biological model is a controlled cigarette-smoke mouse cohort rather than a human COPD cohort. Translation would require clinically feasible predictors, human-relevant muscle endpoints, independent external validation, calibration assessment, subgroup analysis, and prospective evaluation of clinical utility.

The practical implication is that SPD and quantum-kernel methods should be evaluated as targeted representation tools, not as automatic replacements for classical learning. In small biomedical datasets, strong classical baselines should remain the first point of comparison. The present results suggest that when an endpoint is largely linear, as force appeared to be in this cohort, a simple biomarker-only ridge model may be preferable. When an endpoint depends more strongly on covariance structure or nonlinear biomarker relationships, as suggested for muscle weight and muscle quality, RKHS-normalised SPD references and compact quantum-kernel regression can provide additional predictive structure. Future work should test whether the Synth\_ROSE pattern persists in independent preclinical cohorts and in human COPD datasets with clinically feasible measurements. It should also examine whether force prediction requires different biomarkers, within-condition designs, longitudinal measurements, or mechanistic variables to capture functional impairment beyond the dominant exposure-related signal.

\section{Conclusion}

This study provides a leakage-controlled repeated-CV benchmark of classical, SPD-geometric, and simulated quantum methods for predicting skeletal-muscle outcomes in experimental COPD.

The results are endpoint-dependent. Biomarker-only Ridge had the lowest mean RMSE for force readout (2358.459 mN), indicating that this endpoint was best captured by a linear biomarker model in the present cohort. In contrast, Synth\_ROSE had the numerically lowest mean RMSE for muscle quality (58.249 mN/mg) and muscle weight (5.116 mg), with the clearest numerical gain observed for muscle weight relative to the best classical comparator. These gains were numerically favourable but were not statistically significant after paired repeated-CV testing and Holm adjustment, so they should be interpreted as promising endpoint-specific improvements rather than definitive superiority.

Overall, the findings support SPD geometry and RKHS normalisation as useful tools for constructing leakage-safe, low-dimensional representations that can be coupled to compact simulated quantum regression. The study shows that in endpoints with richer biomarker covariance structure, such as muscle weight and muscle quality, geometry-aware quantum-kernel methods can yield lower repeated-CV RMSE than strong classical comparators, while in more linearly structured outcomes, such as force readout, they remain competitive but do not necessarily outperform simpler models. The results do not establish quantum hardware advantage, clinical readiness, or universal superiority over classical learning. Instead, they provide evidence that kernel-geometric quantum representations can offer endpoint-dependent predictive value for biologically meaningful COPD muscle phenotypes and justify further validation in larger preclinical studies and human COPD cohorts.

\end{revblock}

\begin{revblock}
\section*{Supplementary Information}
The Supplementary Information contains the complete unreduced repeated-CV tables, full sensitivity tables, paired repeated-CV statistical tests, method-family descriptions, and the synthetic-medoid Stein--Nystr\"om QKR audit. Main-text rankings use the repeated-CV results from the 15 held-out outer-fold evaluations.

\section*{Data and code availability}
The minimal processed dataset and analysis code are available from the corresponding author upon reasonable request, subject to institutional and ethics requirements.

\section*{Funding}
This work was supported by the National Health and Medical Research Council of Australia under Project Grant Numbers APP1084627 and APP1138915.

\section*{Author contributions}
A.A. conceived the study, developed the computational methodology, performed the formal analysis, implemented the quantum and SPD-based models, prepared the figures, and drafted the manuscript. H.K. contributed to implementation of the machine-learning methods, domain-specific interpretation, figure generation, writing, and editing. S.M.H.C. contributed to biological data collection, study conceptualisation, interpretation of the findings, and critical revision of the manuscript. F.K. contributed to methodological review, development of the quantum machine-learning algorithms, idea generation, and revision of the manuscript. M.U. contributed quantum methodological expertise and revision of the manuscript. R.V. contributed to study supervision, biological interpretation, and critical revision of the manuscript. All authors reviewed and approved the final manuscript.

\section*{Competing interests}
F.K. is affiliated with Pattern Recognition Pty Ltd. The authors declare no competing interests. Pattern Recognition Pty Ltd declares no financial interest in this project.

\section*{Additional information}
Supplementary Information is available for this paper. Correspondence and requests for materials should be addressed to Azadeh Alavi (azadeh.alavi@rmit.edu.au) and Stanley M. H. Chan stanley.chan@rmit.edu.au .

\end{revblock}

\bibliographystyle{naturemag}
\bibliography{references}
\end{document}